\documentclass[runningheads]{llncs}

\usepackage{eccv}

\usepackage{eccvabbrv}

\usepackage{graphicx}
\usepackage{booktabs}
\usepackage{enumitem}
\usepackage{multirow}
\usepackage{placeins}  
\usepackage{microtype} 
\usepackage{colortbl}  
\usepackage{tabularx}  
\usepackage{makecell}  
\newcolumntype{Y}{>{\centering\arraybackslash}X}  
\newcommand{\hd}[2]{\begin{tabular}[b]{@{}c@{}}#1\\#2\end{tabular}}  
\definecolor{cellbest}{rgb}{0.99,0.74,0.71}    
\definecolor{cellsecond}{rgb}{1.0,0.92,0.66}   
\renewcommand{\tablename}{Tab.}                

\usepackage[accsupp]{axessibility}  

\newcommand{\methodname}{OVOW\xspace}

\usepackage{hyperref}

\usepackage{orcidlink}

\begin{document}

\title{\underline{O}ne \underline{V}ideo, \underline{O}ne \underline{W}orld: Turning Monocular Video into Physical 4D Scenes}

\titlerunning{One Video, One World}

\author{
Junhao Chen\inst{1,2}\textsuperscript{$\star$}\orcidlink{0009-0006-4195-3766} \and
Boran Zhang\inst{3}\textsuperscript{$\star$}\orcidlink{0009-0003-0708-1659} \and
Mingjin Chen\inst{4}\orcidlink{0009-0005-4979-6307} \and
Henghaofan Zhang\inst{5}\orcidlink{0009-0003-2714-2092} \and \\
Saining Zhang\inst{6}\orcidlink{0009-0000-4983-8478} \and
Congcong Zhu\inst{3}\orcidlink{0000-0001-5146-222X} \and
Hao Zhao\inst{7}\orcidlink{0000-0001-7903-581X} \and
Ruqi Huang\inst{1}\textsuperscript{$\dagger$}\orcidlink{0000-0001-5942-3671} \and \\
Zhihao Li\inst{2}\orcidlink{0000-0002-2066-8775} \and
Yufei Wang\inst{2}\textsuperscript{$\dagger\S$}\orcidlink{0000-0002-6326-7357}
}

\authorrunning{J.~Chen et al.}

\institute{
Shenzhen International Graduate School, Tsinghua University, China
\email{dreamhowchen@gmail.com, ruqihuang@sz.tsinghua.edu.cn} \and
SparcAI Inc, USA \\
\email{yufei.wang@sparclab.ai} \and
University of Science and Technology of China, China \and
The Hong Kong Polytechnic University, China \and
University of Electronic Science and Technology of China, China \and
Nanyang Technological University, Singapore \and
Institute for AI Industry Research (AIR), Tsinghua University, China \\
}

\maketitle

\renewcommand{\thefootnote}{}
\footnotetext{\textsuperscript{$\star$}Equal Contribution. Work performed during an internship. \textsuperscript{$\dagger$}Corresponding Author. \textsuperscript{$\S$}Project Lead. 
Project Page at \url{https://OneVideoOneWorld.github.io}.
}
\setcounter{footnote}{0}
\renewcommand{\thefootnote}{\arabic{footnote}}

\begin{center}
    \vspace{-3mm}
    Project Page: \url{https://OneVideoOneWorld.github.io} \\
    \vspace{-5mm}
\end{center}

 \begin{figure*}[!ht]
  \centering
  \includegraphics[width=\textwidth]{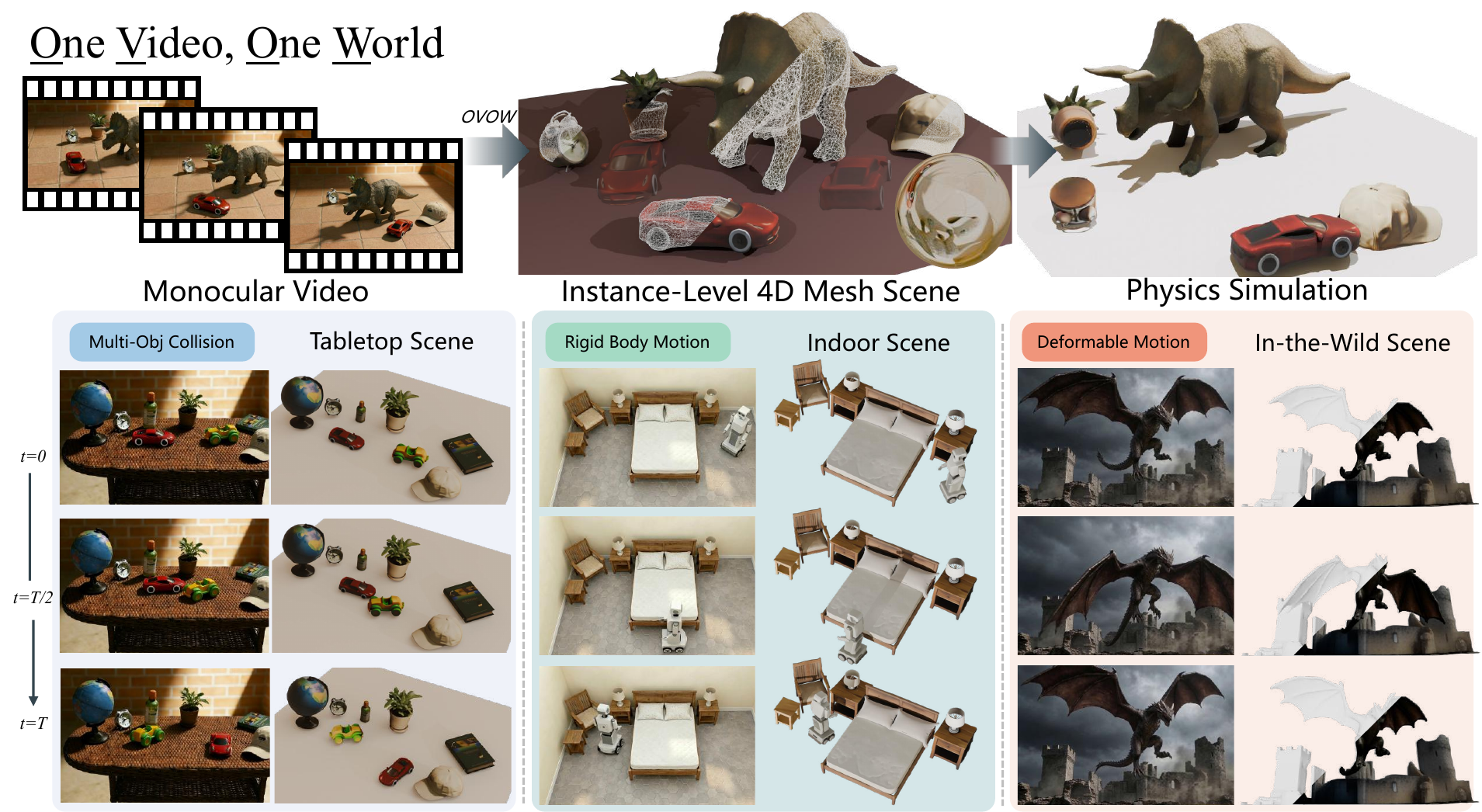}
\caption{\textbf{OVOW reconstructs instance-level, simulation-ready 4D mesh scenes from monocular video.} Given a single video, our method decomposes the scene into physically independent mesh instances and recovers rigid-body motions and non-rigid mesh deformations, yielding instance-level meshes ready for downstream physics simulation and editing. We demonstrate results of multi-object collisions, rigid-body motions, and deforming object motions across tabletop, indoor, and in-the-wild scenarios.}
  \label{fig:teaser}
\end{figure*}

\begin{abstract}
  We introduce \textbf{OVOW}, the first training-free system that reconstructs \emph{instance-level, simulation-ready} 4D mesh scenes from a single monocular video.
Recent 4D reconstruction achieves impressive rendering quality, but its outputs (\eg, implicit fields, Gaussian primitives, or point clouds) lack the watertight topology, instance separation, and standardized physical interfaces required by physics simulators and embodied AI.
OVOW closes this gap with a four-stage pipeline: a vision-language model discovers, labels, and motion-classifies all instances; category-aware reconstruction yields per-instance meshes for rigid objects and topology-consistent mesh sequences for deformable ones; an iterative render-match-optimize procedure recovers metric scale and 6-DoF pose trajectories; and physics-grounded assembly enforces ground contact and inter-object support.
Crucially, we model all motion, rigid and non-rigid, through direct vertex deformation without category-specific priors or skeleton rigging, producing watertight mesh scenes ready for downstream physics simulation and editing.
We further establish the first benchmark for \emph{structured Video-to-4D} evaluation, with metrics for geometric correctness, instance separation, and physical plausibility beyond visual fidelity; the same pipeline doubles as a scalable engine for \emph{synthesizing} paired video-to-4D simulation data for future 4D world models and embodied AI.
Across two synthetic benchmarks (static and 4D), OVOW attains the best overall layout and geometry accuracy and the lowest photometric and semantic error among all baselines, and on monocular video runs one to two orders of magnitude faster than the baselines, while downstream physics simulation confirms its physical stability.

  \keywords{Simulation-Ready Asset Generation \and Video-to-4D \and Instance-Level Scene Reconstruction}
\end{abstract}

\section{Introduction}
\label{sec:intro}

Recent advances in 4D reconstruction~\cite{dnerf,nerfies,4dgs,sv4d,l4gm} have dramatically improved the visual quality of dynamic scene rendering, yet a fundamental disconnect persists: \textbf{no existing method can produce simulation-ready 4D scene assets from video}.
Physics simulators that underpin modern robotics and embodied AI, such as MuJoCo~\cite{mujoco}, Isaac Gym~\cite{isaacgym}, and PyBullet~\cite{pybullet}, require watertight meshes, physically separated instances, and standardized interfaces such as URDF\@, which are absent from the implicit fields, Gaussian primitives, and point clouds produced by current methods~\cite{nerf,3dgs,4dgs}.
The field excels at \emph{looking real} but cannot yet produce \emph{physically usable} outputs for robotic manipulation, embodied AI, gaming, and virtual reality.

Two converging trends suggest this gap can now be closed.
On the \emph{generation} side, agent-driven scene synthesis~\cite{viga,code2worlds,sage,scenethesis,tabletopgen,pat3d,zhu2025imaginarium} shows growing demand for simulation-ready, instance-level assets~\cite{awesome3dscenegen}, validating our target format; on the \emph{reconstruction} side, mesh-oriented 4D~\cite{mesh4d,dgmesh,motion324} and scene-level 3D~\cite{midi,cast,sencemaker} methods show that temporally consistent mesh recovery from monocular video is feasible at the single-object level~\cite{4dressurvey,4Dadvances,kong20253d}.
What remains missing is a \textbf{unified pipeline that scales from individual objects to entire scenes}, recovering instance-level meshes with rigid and non-rigid motion and assembling them into a physically coherent, simulation-ready representation.

Achieving this requires addressing two challenges.
The \textbf{first} is scene representation: how to convert a dynamic multi-object scene into simulation-compatible structured assets.
Unlike skeleton-based approaches~\cite{SMPL2015,unirig,riganything}, which require predefined joint topologies and suffer from skinning artifacts on complex deformations, we adopt \emph{direct mesh vertex deformation} as a unified motion model: each object is a mesh described by rigid-body transformations (static/rigid) or per-vertex displacement fields (deformable).
This needs no predefined kinematic chains, handles articulated motion, surface deformation, and soft-body dynamics, makes \emph{no assumptions} about object category or topology, and yields watertight, instance-level meshes ready for downstream physics simulation and editing.

The \textbf{second} challenge is data and evaluation.
No dataset pairs instance-level 4D mesh annotations with source video, and no benchmark measures the structural qualities that matter for downstream physical tasks, such as geometric correctness, instance separation, and physical interaction, rather than rendering fidelity alone (PSNR/SSIM/LPIPS).
We construct the \emph{first benchmark for structured Video-to-4D evaluation}, assessing geometric and physical qualities beyond visual fidelity on carefully designed synthetic data, and further use our scalable pipeline to generate ``video $\leftrightarrow$ instance-level 4D mesh scene'' pairs as data infrastructure for future 4D world-model research and embodied AI.

With these components, we present \textbf{\methodname} (Fig.~\ref{fig:teaser}), the first method to reconstruct instance-level, simulation-ready 4D mesh scenes from monocular video.
\methodname~is \emph{fully training-free}, composing pre-trained foundation models for scene understanding, mesh reconstruction, and metric recovery into a tightly integrated pipeline, and performs robustly across diverse in-the-wild scenarios spanning indoor, outdoor, and complex multi-object scenes with rigid and non-rigid motion.
On our two synthetic benchmarks for static and structured-4D scenes, \methodname~achieves the best overall layout and geometry accuracy and the lowest photometric and semantic error among all baselines, and on monocular video runs one to two orders of magnitude faster than the baselines.
Downstream physics simulation confirms that the reconstructed scenes remain physically stable.
Our contributions are three-fold:
\begin{enumerate}[leftmargin=1.5em,itemsep=1pt]
    \item \textbf{Task \& method.} We formalize \emph{structured Video-to-4D reconstruction} and present the first training-free pipeline that turns a monocular video into instance-level, simulation-ready 4D mesh scenes, modeling \emph{all} motion via direct vertex deformation without category priors or skeleton rigging.
    \item \textbf{Benchmark.} We build the first benchmark for the task, scoring geometric correctness, instance separation, and physical plausibility beyond visual fidelity.
    \item \textbf{Data engines.} We contribute two complementary data pipelines: \methodname turns real video into paired ``video$\leftrightarrow$4D scene'' data, and an asset-based pipeline composes the synthetic 3D/4D scenes of our benchmark; together they supply the simulation-ready paired supervision this task lacks.
\end{enumerate}

\section{Related Work}
\label{sec:related}

\subsection{4D Reconstruction and Generation}
\label{sec:rw_4d}
3D and 4D reconstruction and generation have advanced rapidly across perception, generation, and single-object dynamics. Monocular depth estimation~\cite{depthv1,depthv2,depthv3,miao2026framessequencestemporallyconsistent} and feed-forward 3D prediction~\cite{dust3r,monst3r,vggt} recover geometry as unstructured point clouds; neural scene representations have evolved from static fields and Gaussians to dynamic 4D~\cite{hypernerf,nsff,kplanes,hexplane,zhong2026tidegs}; image/text-driven 3D generation~\cite{wonder3d,idea23d,xiang2024structured,chen2026ultraman,xiang2025native,lisparc3d,zhang2024clay,weng2026partflow} has extended to 4D via video-diffusion priors and feed-forward models~\cite{comp4d,diffusion4d,cao2025physx,liu2025free4d,lin2025phys4dgen,singer2023text,li2024dreammesh4d,gengone}; and mesh-oriented~\cite{tessgs,chen2025v2m4,sabathier2026actionmesh} and motion-recovery~\cite{casa,magicpose4d,riggs,wu2025animateanymesh} methods recover single-object dynamic geometry. Yet these outputs remain rendering-oriented and largely single-object, lacking the watertight topology and instance separation~\cite{4dreview1,4dreview2} that physics requires. \methodname instead recovers watertight, instance-separated meshes for complete dynamic scenes, rather than single-object or rendering-only outputs.

\subsection{Structured Scene Generation}
\label{sec:rw_scene}
A parallel line targets structured, multi-object 3D scenes, building on structured generation across text~\cite{iwbench,llmspark,zhujiu}, image/video~\cite{ye2024mmad,lottiegpt}, and 3D~\cite{weng2026garmentgpt,wang2024llama}. Instance-level reconstruction~\cite{sam3d,zhao2025depr,yang2025instascene,siddiqui2026shaper} recovers component-separated scenes from images, compositional generators~\cite{han2025reparo,chengraph2scene,hu2026mixed,tang2025zeroscene,mengscenegen,chen2025layout2scene3dsemanticlayout,wang2026scenetransporter,Chen_2026_CVPR_CustomTex} assemble multi-object scenes, and feed-forward~\cite{karhade2025any4d,flux4d,split4d} and instance-aware~\cite{dreamscene4d} methods reconstruct multi-object motion as point clouds or Gaussians. Crucially, these build scenes from text or procedural rules, \emph{not} real video; reconstructing structured scenes from video would unlock vast internet and robot footage~\cite{chen2026dancetogether,chen2026hvg3dbridgingrealsimulation}, yet existing 4D datasets~\cite{objaverse,wu2025animateanymesh} and benchmarks~\cite{charge,zhu20254d} lack source-video-paired instance-level annotations and structural metrics. In contrast, \methodname reconstructs instance-level scenes directly from real monocular video and supplies the paired data and structural benchmark this research line lacks.

\subsection{Simulation-Ready Asset Generation}
\label{sec:rw_v24d}
A growing body of work pursues assets that are ready for physical simulation. Automatic rigging~\cite{magicarticulate,xu2020rignet,sun2025drive,guo2025auto,sun2026animatorcentricskeletongenerationobjects,guo2025make} and articulated-object generation~\cite{articulateanymesh,song2025magicarticulate,song2025puppeteer} endow individual assets with kinematic structure such as joints and hinges for part-level articulation, while agent-driven scene generators~\cite{artiscene,embodiedgen,ling2025scene} compose whole scenes that explicitly target simulation readiness. Together, these reflect the growing demand from physics simulators~\cite{sapien} and embodied world models~\cite{tesseract} for mesh-based, physically interactive assets. However, all of these methods build assets from synthetic priors, category templates, or manual design rather than real-world visual observation. \methodname is the first to recover instance-level, simulation-ready scene meshes with both rigid and non-rigid motion directly from a monocular video, without per-category rigging or manual articulation.

\begin{figure*}[!t]
    \centering
    \includegraphics[width=1\linewidth]{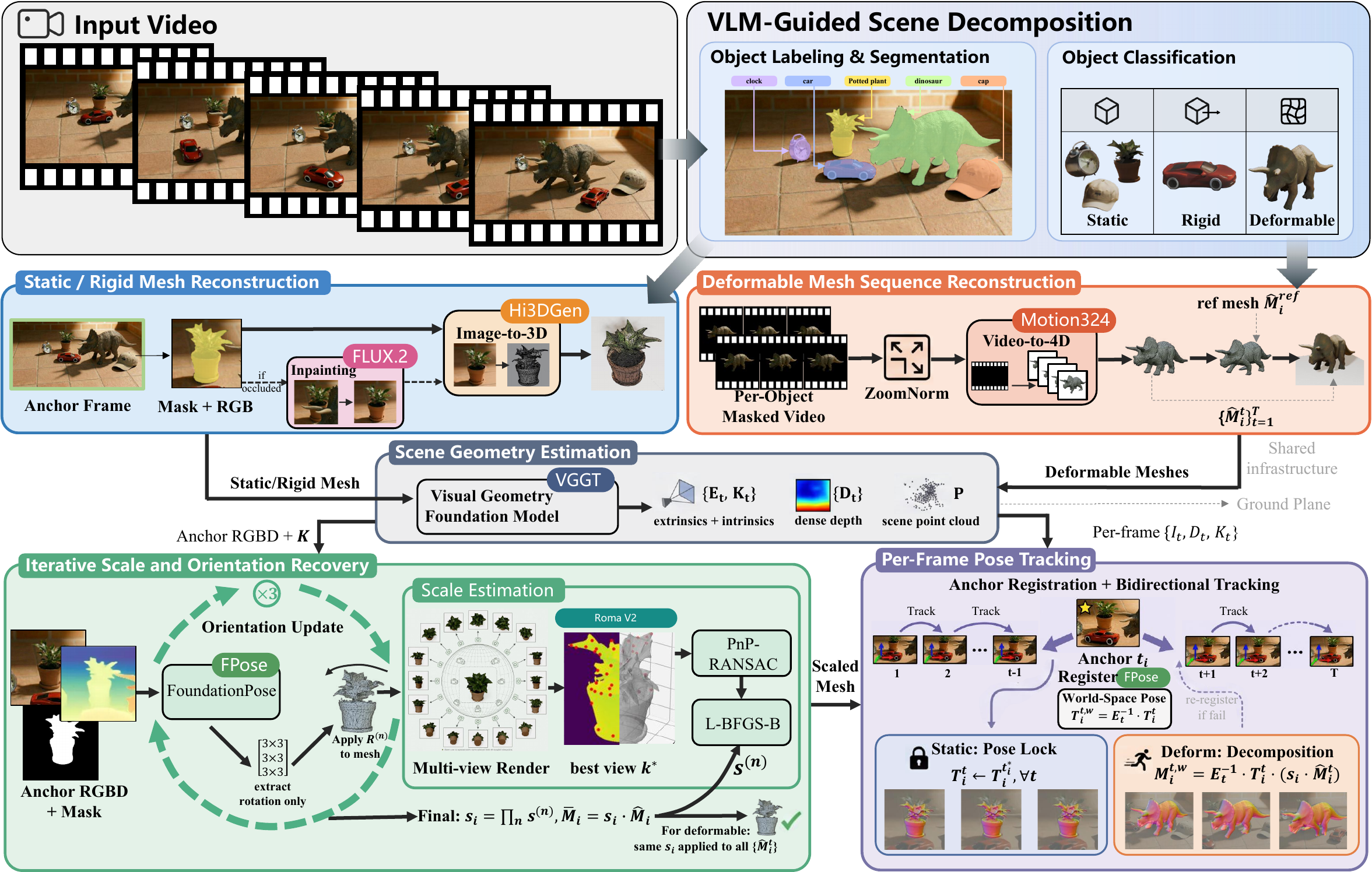}
    \caption{\textbf{Overview of \methodname{} (Stages~1--3).} From a single monocular video, our training-free pipeline decomposes the scene into labeled, motion-classified instances, reconstructs per-instance static/rigid meshes and deformable mesh sequences, and recovers metric scale and per-frame 6-DoF motion; the recovered instances are then assembled in Fig.~\ref{fig:pipe_sim}. Each stage is detailed in Sections~\ref{sec:scene_understanding}--\ref{sec:pose_motion}.}
    \label{fig:pipe_track}
\end{figure*}

\section{Method}
\label{sec:method}

Given a monocular video $\mathcal{V} = \{I_t\}_{t=1}^{T}$ depicting a dynamic multi-object scene, our goal is to reconstruct a \emph{simulation-ready} 4D scene
\begin{equation}
    \mathcal{S} = \bigl\{(\mathcal{M}_i,\; \{\mathbf{T}_i^t\}_{t=1}^{T})\bigr\}_{i=1}^{N},
    \label{eq:scene_def}
\end{equation}
where each of the $N$ objects is represented by a watertight triangle mesh $\mathcal{M}_i = (\mathbf{V}_i, \mathbf{F}_i)$ with vertices $\mathbf{V}_i \in \mathbb{R}^{|\mathbf{V}_i|\times 3}$ and faces $\mathbf{F}_i$, together with a per-frame 6-DoF pose trajectory $\mathbf{T}_i^t \in \mathrm{SE}(3)$.
For deformable objects, $\mathcal{M}_i$ is replaced by a topology-consistent mesh sequence $\{\mathcal{M}_i^t\}_{t=1}^{T}$ with per-vertex displacement fields.
The complete scene is output as instance-level, simulation-ready meshes for downstream physics simulation and editing.

\begin{figure*}[!t]
    \centering
    \includegraphics[width=1\linewidth]{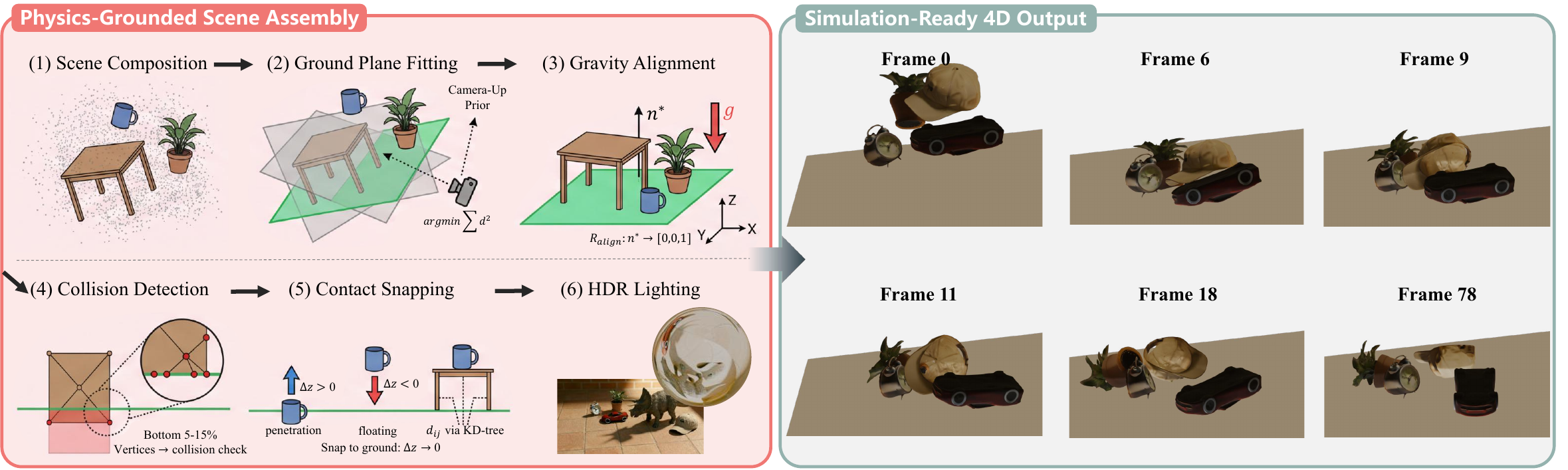}
    \caption{\textbf{Physics-Grounded Scene Assembly and Simulation-Ready Output (Stage~4).} The instances from Fig.~\ref{fig:pipe_track} are assembled into a physically coherent scene by ground-plane estimation and contact projection, then exported in URDF format for downstream physics simulation (Section~\ref{sec:physics}).}
    \label{fig:pipe_sim}
\end{figure*}

As illustrated in Figs.~\ref{fig:pipe_track} and~\ref{fig:pipe_sim}, our pipeline operates in a \textbf{fully training-free} manner and comprises four stages:
\textbf{(1)}~\emph{VLM-Guided Scene Decomposition} (\S\ref{sec:scene_understanding});
\textbf{(2)}~\emph{Instance-Level Mesh Reconstruction} (\S\ref{sec:mesh_recon});
\textbf{(3)}~\emph{Spatiotemporal Pose and Deformation Recovery} (\S\ref{sec:pose_motion}); and
\textbf{(4)}~\emph{Physics-Grounded Scene Assembly} (\S\ref{sec:physics}).

\subsection{VLM-Guided Scene Decomposition}
\label{sec:scene_understanding}

We uniformly sample $N_{\text{key}}$ keyframes (default $N_{\text{key}}{=}3$) from the input video and encode them into a multi-image prompt for a VLM~\cite{qwen3vl}.
The model jointly performs \emph{open-vocabulary object discovery}, \emph{unique instance naming} (in the form \texttt{<noun>\_<index>}), and \emph{motion category classification} into one of three categories: \textbf{static} (unchanged pose), \textbf{rigid} (rigid-body motion), or \textbf{deformable} (non-rigid deformation).
The output is a structured JSON with $N$ records $\{(\ell_i, c_i, d_i)\}_{i=1}^{N}$, where $\ell_i$ is the unique label, $c_i \in \{\text{static}, \text{rigid}, \text{deform}\}$ is the motion category, and $d_i$ is a brief visual description.

Given the label set $\{\ell_i\}$, we perform dense video segmentation using SAM3~\cite{sam3} with text prompts, producing per-frame binary masks
\begin{equation}
    \{M_i^t \in \{0,1\}^{H \times W}\}_{t=1}^{T}, \quad i = 1, \ldots, N.
    \label{eq:masks}
\end{equation}
Instances with maximum mask area below $\tau_{\text{area}}$ (default 200\,px) are discarded.

\subsection{Instance-Level Mesh Reconstruction}
\label{sec:mesh_recon}

\subsubsection{Static and Rigid Object Mesh Generation}
\label{sec:mesh_static_rigid}

For each static or rigid object $i$ ($c_i \in \{\text{static}, \text{rigid}\}$), we select the \emph{anchor frame} $t_i^* = \arg\max_t \|M_i^t\|_1$ with the largest mask area.
When the object is partially occluded, we apply an inpainting model~\cite{flux2} conditioned on the masked RGB, the mask, and the VLM description $d_i$ to hallucinate the complete appearance.
The completed image is then fed to a feed-forward image-to-3D model~\cite{ye2025hi3dgen}, producing a canonical mesh $\hat{\mathcal{M}}_i$.

\subsubsection{Deformable Object Mesh Sequence Reconstruction}
\label{sec:mesh_deform}

For each deformable object $i$ ($c_i = \text{deform}$), we extract a per-object masked video and apply \emph{zoom normalization} to stabilize the object's position and scale across frames:
\begin{equation}
    \hat{I}_i^t = \text{ZoomNorm}\bigl(I_t \odot M_i^t,\; \text{BBox}(M_i^t),\; \rho,\; L\bigr),
    \label{eq:zoom_norm}
\end{equation}
where $\odot$ denotes element-wise masking, $\rho$ is the target object-to-frame ratio (default $0.65$), and $L$ is the output resolution (default $512$).
The normalized video is processed by a deformable mesh reconstruction model~\cite{motion324}, which outputs a topology-consistent mesh sequence $\{\hat{\mathcal{M}}_i^t\}_{t=1}^{T}$ and a reference mesh $\hat{\mathcal{M}}_i^{\text{ref}}$.

\subsubsection{Metric Scale Recovery}
\label{sec:scale}

We employ a visual geometry foundation model~\cite{vggt} to obtain per-frame camera extrinsics $\{E_t \in \mathrm{SE}(3)\}_{t=1}^{T}$, intrinsics $\{K_t\}_{t=1}^{T}$, dense depth maps $\{D_t\}_{t=1}^{T}$, and a scene-level point cloud $\mathcal{P} \in \mathbb{R}^{P \times 3}$.
For each object $i$, at the anchor frame $t_i^*$, we back-project the masked depth into 3D:
\begin{equation}
    \mathcal{Q}_i = \Pi^{-1}\bigl(M_i^{t_i^*} \odot D_{t_i^*},\; K_{t_i^*}\bigr) \in \mathbb{R}^{Q_i \times 3},
    \label{eq:backproject}
\end{equation}
and compute an initial scale factor $s_i^{(0)}$ by aligning bounding-box diagonals:
\begin{equation}
    s_i^{(0)} = \frac{\bigl\|\text{diag}(\mathcal{Q}_i)\bigr\|}{\bigl\|\text{diag}(\hat{\mathcal{M}}_i)\bigr\|}.
    \label{eq:scale}
\end{equation}
This estimate is refined into the final scale $s_i$ (\S\ref{sec:scale_orient}); the scaled mesh $\bar{\mathcal{M}}_i = s_i \cdot \hat{\mathcal{M}}_i$ is used in all subsequent stages.
For deformable objects, the same $s_i$ is applied uniformly to all mesh frames.

\subsection{Spatiotemporal Pose and Deformation Recovery}
\label{sec:pose_motion}

We adopt a unified two-stage procedure for all object categories: \emph{iterative scale-orientation recovery} at the anchor frame, followed by \emph{bidirectional pose tracking} across all frames.
For deformable objects, the per-frame mesh sequence replaces the single canonical mesh; otherwise the pipeline is identical.

\subsubsection{Iterative Scale and Orientation Recovery}
\label{sec:scale_orient}

For each object $i$, we jointly refine the metric scale $s_i$ and orientation on the anchor frame $t_i^*$ through a \emph{render-match-optimize} loop, iterated $N_{\text{iter}}$ times (default $N_{\text{iter}}{=}3$).
At each iteration $n$:

\noindent\textbf{(a) Orientation Update.}
FoundationPose~\cite{foundationpose} estimates a 6-DoF pose from the anchor-frame RGBD $(I_{t_i^*}, D_{t_i^*})$, mask $M_i^{t_i^*}$, and intrinsic $K_{t_i^*}$; we extract only the rotation $R^{(n)}$ and apply it to the mesh.

\noindent\textbf{(b) Scale Estimation.}
The rotated mesh is rendered from $J$ viewpoints on a sphere with z-buffer depth $\{\tilde{D}_k\}$.
We establish dense correspondences between each rendered view and the real masked crop using the dense feature matcher RoMa~v2~\cite{edstedt2025roma}. The view $k^*$ with the most confident matches is selected, yielding 2D-3D correspondences:
\begin{equation}
    \mathbf{q}_m^{\text{mesh}} = \Pi^{-1}(\mathbf{p}_m^{\text{render}},\, \tilde{D}_{k^*},\, K_{t_i^*}), \quad
    \mathbf{q}_m^{\text{depth}} = {R^{(n)}}^\top \cdot \Pi^{-1}(\mathbf{p}_m^{\text{real}},\, D_{t_i^*},\, K_{t_i^*}).
    \label{eq:corr_3d}
\end{equation}
A pose is obtained via PnP-RANSAC~\cite{fischler1981random,lepetit2009ep}, and the scalar scale $s^{(n)}$ is optimized via L-BFGS-B~\cite{byrd1995limited} after centering both point sets:
\begin{equation}
    s^{(n)} = \arg\min_{s} \sum_{m} w_m \, \bigl\|\bar{\mathbf{q}}_m^{\text{depth}} - s \cdot \bar{\mathbf{q}}_m^{\text{mesh}}\bigr\|_2^2,
    \label{eq:scale_lbgfs}
\end{equation}
where $\bar{\mathbf{q}}$ denotes centered coordinates, $w_m$ is the match confidence, and the top $5\%$ residuals are trimmed.
The final scale is $s_i = \prod_{n} s^{(n)}$ (with $s^{(0)}{=}s_i^{(0)}$ the initial estimate), yielding $\bar{\mathcal{M}}_i = s_i \cdot \hat{\mathcal{M}}_i$.

\subsubsection{Per-Frame Pose Tracking}
\label{sec:pose_track}

With $\bar{\mathcal{M}}_i$, we estimate $\mathbf{T}_i^t \in \mathrm{SE}(3)$ at every frame via FoundationPose~\cite{foundationpose}:
\begin{equation}
    \mathbf{T}_i^t = \arg\min_{\mathbf{T}} \; \mathcal{L}_{\text{render}}\bigl(\text{Render}(\bar{\mathcal{M}}_i, \mathbf{T}, K_t),\; I_t, D_t, M_i^t\bigr).
    \label{eq:fpose}
\end{equation}
We apply \emph{mask-gated inputs} (zeroing RGB/depth outside $M_i^t$) to focus registration on the target instance.
At the anchor frame $t_i^*$ the estimator performs full registration; from there we track bidirectionally:
\begin{equation}
    \mathbf{T}_i^{t \pm 1} = \texttt{Track}\bigl(\mathbf{T}_i^{t},\; I_{t \pm 1},\; D_{t \pm 1},\; K_{t \pm 1}\bigr),
    \label{eq:track}
\end{equation}
falling back to full re-registration when tracking fails.
The world-space pose is:
\begin{equation}
    \mathbf{T}_i^{t,\text{w}} = E_t^{-1} \cdot \mathbf{T}_i^t.
    \label{eq:world_pose}
\end{equation}

\noindent\textbf{Static Pose Locking.}
For \emph{static} objects, we lock all frames to the pose at the anchor frame $t_i^*$ (the largest-mask frame):
\begin{equation}
    \mathbf{T}_i^t \leftarrow \mathbf{T}_i^{t_i^*}, \quad \forall\, t \neq t_i^*.
    \label{eq:lock_static}
\end{equation}

\noindent\textbf{Deformable Object Decomposition.}
For deformable objects, the world-space mesh at frame $t$ is:
\begin{equation}
    \mathcal{M}_i^{t,\text{w}} = \mathbf{T}_i^{t,\text{w}} \cdot \bar{\mathcal{M}}_i^t = E_t^{-1} \cdot \mathbf{T}_i^t \cdot (s_i \cdot \hat{\mathcal{M}}_i^t),
    \label{eq:deform_world}
\end{equation}
decoupling global rigid trajectory from local non-rigid deformation.

\subsection{Physics-Grounded Scene Assembly}
\label{sec:physics}

The independently reconstructed objects must be assembled into a physically coherent scene. We achieve this through ground plane estimation followed by contact projection.

\noindent\textbf{Ground Plane Estimation.}
We extract $N_{\text{plane}}$ candidate planes $\{(\mathbf{n}_j, d_j)\}$ (default $5$) from $\mathcal{P}$ via iterative RANSAC, and select the one that jointly maximizes the \emph{above-plane ratio} $r_j = |\{\mathbf{v}: \mathbf{v}^\top \mathbf{n}_j + d_j \geq -\epsilon\}|/|\mathbf{V}_{\text{static}}|$ and minimizes the \emph{contact proximity} $\tilde{d}_j = \text{median}(\text{bottom}_{p\%}(|\mathbf{v}^\top \mathbf{n}_j + d_j|))$ (default $p{=}3$).
A \emph{camera-up prior} resolves normal sign ambiguity:
\begin{equation}
    \text{if } \text{median}_t \bigl((\mathbf{n}_j^\top R_t)_y\bigr) > 0, \quad \text{flip: } \mathbf{n}_j \leftarrow -\mathbf{n}_j, \; d_j \leftarrow -d_j,
    \label{eq:camera_up}
\end{equation}
where $R_t$ is the rotation block of $E_t$.
The selected plane is aligned to the XY-plane via $R_{\text{align}}$ mapping $\mathbf{n}^* \!\to\! [0,0,1]^\top$.

\noindent\textbf{Contact Projection.}
Objects are sorted by bottom distance to the ground per frame; for each object $i$, we extract the bottom $p_{\text{bot}}\%$ (default $10\%$) vertices by signed distance $d_{\mathbf{n}}(\mathbf{v}) = \mathbf{v}^\top \mathbf{n}^* + d^*$:
\begin{equation}
    \mathcal{B}_i^t = \bigl\{\mathbf{v} \in \mathbf{V}_i^{t,\text{w}}\!: d_{\mathbf{n}}(\mathbf{v}) \leq \text{quantile}_{p_{\text{bot}}}(\{d_{\mathbf{n}}(\mathbf{v}')\}_{\mathbf{v}'})\bigr\}.
    \label{eq:bottom_set}
\end{equation}
The gravity-axis correction is:
\begin{equation}
    \Delta z_i^t = \begin{cases}
    -(d_{\min} + \epsilon), & d_{\min} < -\epsilon \;\text{(penetration)}, \\[3pt]
    -\min(d_{\min} - \epsilon,\; \delta_{\max}), & d_{\min} > \epsilon \;\text{(floating)}, \\[3pt]
    0, & \text{otherwise},
    \end{cases}
    \label{eq:z_correction}
\end{equation}
where $d_{\min} = \min_{\mathbf{v} \in \mathcal{B}_i^t} d_{\mathbf{n}}(\mathbf{v})$, $\epsilon$ is a contact tolerance ($0.2\%$ of scene diagonal), and $\delta_{\max}$ caps downward settling ($2\%$).

\noindent\textbf{Inter-Object Contact.}
We query the nearest-surface distance from $\mathcal{B}_i^t$ to each settled object $j$ via KD-tree:
\begin{equation}
    d_{ij}(\mathbf{v}) = \min_{\mathbf{v}' \in \mathbf{V}_j^{t,\text{w}}} \|\mathbf{v} - \mathbf{v}'\|_2, \quad \mathbf{v} \in \mathcal{B}_i^t.
    \label{eq:contact_dist}
\end{equation}
If $d_{ij} \leq \tau_{\text{contact}}$ ($3\%$ of scene diagonal), the surface of $j$ serves as a local ground plane for $i$, naturally handling stacking configurations.
The procedure is iterated $N_{\text{asm}}$ times (default $2$) per frame to resolve cascading dependencies.

\noindent\textbf{Environment Lighting Recovery.}
To enable photorealistic rendering of the assembled scene, we recover an HDR environment map from the input video using an intrinsic decomposition model~\cite{dille2024intrinsic}, which is applied as world lighting in the exported scene.

\section{Experiments}
\label{sec:results}

\subsection{Implementation Details}
\label{sec:impl}
\methodname composes off-the-shelf foundation models without task-specific training: Qwen3-VL~\cite{qwen3vl} and SAM3~\cite{sam3} for decomposition and segmentation; FLUX.2~\cite{flux2} amodal inpainting with the feed-forward generator Hi3DGen~\cite{ye2025hi3dgen} for static/rigid meshes; Motion324~\cite{motion324} for deformable meshes; VGGT~\cite{vggt} for scene geometry; RoMa~v2~\cite{edstedt2025roma} for dense correspondences; and FoundationPose~\cite{foundationpose} for 6-DoF tracking.
Default settings: $N_{\text{key}}{=}3$ keyframes, $\tau_{\text{area}}{=}200$\,px, $N_{\text{iter}}{=}3$, $N_{\text{asm}}{=}2$, $\rho{=}0.65$, $L{=}512$, contact tolerance $0.2\%$, settling cap $2\%$, and inter-object contact threshold $3\%$ of the scene diagonal.

\subsection{Benchmark Setup}
\label{sec:benchmark}

\noindent\textbf{Evaluation Dataset.}
Separate from \methodname (which reconstructs 4D scenes from real video), we render the evaluation benchmark with a synthetic pipeline that \emph{composes} scenes from existing 3D objects and HDRI assets in Blender, split into two complementary subsets: \textbf{OVOW-3D-Scene-Bench}, with \textbf{120} fully static scenes, and \textbf{OVOW-4D-Scene-Bench}, with \textbf{120} dynamic scenes in which at least one object undergoes rigid-body motion while the rest stay static.
We sample 3D instances from Objaverse-OA~\cite{lu2025orientation} and compose a random \textbf{3--5} of them per scene on a ground plane with Poly Haven\footnote{\url{https://github.com/Poly-Haven/polyhavenassets}} HDRI lighting and backgrounds, animating the moving objects along procedural 6-DoF trajectories for reproducible dynamics.
Every scene yields ground-truth instance meshes, 6-DoF pose trajectories, and camera parameters; full construction details are in the appendix.

\noindent\textbf{Baselines.}
Since no existing method directly addresses instance-level 4D scene reconstruction from video, we compare against state-of-the-art single-image scene reconstruction methods: CAST~\cite{cast}, SAM3D~\cite{sam3d}, VIGA~\cite{viga}, MIDI~\cite{midi}, SceneGen~\cite{mengscenegen}, and TabletopGen~\cite{tabletopgen}.
All of these baselines support only single-frame image input; even on OVOW-4D-Scene-Bench they cannot consume video, so we run each independently on seven uniformly sampled frames per scene and average the resulting scores (runtime and peak VRAM are reported per generated frame), whereas \methodname consumes the full video.

\noindent\textbf{Metrics.}
We report three IoU measures: \emph{Scene-IoU} is the volumetric IoU of the predicted vs.\ ground-truth unions of object boxes, under axis-aligned (AABB$\uparrow$) and oriented (OBB$\uparrow$) boxes for global layout, and \emph{Object-IoU}$\uparrow$ is the per-object IoU after Hungarian matching.
We also report photometric loss (PL$\downarrow$, $100{\times}$ normalized RGB MSE), negative-CLIP score (N-CLIP$\downarrow$, $10{\times}(1{-}\text{CLIP sim})$), both lower-is-better, plus per-frame time and peak VRAM.

\begin{table}[t]
    \centering
    \scriptsize
    \setlength{\tabcolsep}{2.5pt}
    \renewcommand{\arraystretch}{1.05}
    \caption{Quantitative comparison on \textbf{OVOW-3D-Scene-Bench}.}
    \label{tab:quant_static}
    \begin{tabularx}{\linewidth}{@{}l*{7}{Y}@{}}
        \toprule
        Method & \hd{Scene-IoU}{AABB\,$\uparrow$} & \hd{Scene-IoU}{OBB\,$\uparrow$} & \hd{Object}{IoU\,$\uparrow$} & \hd{PL}{$\downarrow$} & \hd{N-CLIP}{$\downarrow$} & \hd{Time}{{\tiny(s/frame)}\,$\downarrow$} & \hd{VRAM}{{\tiny(GB)}\,$\downarrow$} \\
        \midrule
        CAST~\cite{cast}              & 0.077 & 0.178 & 0.097 & \cellcolor{cellsecond}7.90 & 2.02 & 365 & \cellcolor{cellbest}\textbf{10.5} \\
        SAM3D~\cite{sam3d}            & 0.062 & 0.151 & 0.060 & 9.80 & 2.08 & \cellcolor{cellbest}\textbf{103} & 33.5 \\
        VIGA~\cite{viga}              & \cellcolor{cellbest}\textbf{0.156} & 0.158 & 0.024 & 24.40 & 2.61 & 788 & 31.5 \\
        MIDI~\cite{midi}              & 0.092 & \cellcolor{cellsecond}0.216 & \cellcolor{cellsecond}0.136 & 8.80 & 1.91 & 234 & 17.7 \\
        SceneGen~\cite{mengscenegen}  & 0.098 & 0.207 & 0.078 & 9.30 & \cellcolor{cellsecond}1.89 & \cellcolor{cellsecond}152 & 29.6 \\
        TabletopGen~\cite{tabletopgen}& 0.066 & 0.155 & 0.035 & 18.20 & 2.17 & 638 & \cellcolor{cellsecond}14.0 \\
        \midrule
        \textbf{OVOW (Ours)}          & \cellcolor{cellsecond}0.130 & \cellcolor{cellbest}\textbf{0.218} & \cellcolor{cellbest}\textbf{0.190} & \cellcolor{cellbest}\textbf{5.70} & \cellcolor{cellbest}\textbf{1.87} & 272 & 26.0 \\
        \bottomrule
    \end{tabularx}
\end{table}

\begin{table}[t]
    \centering
    \scriptsize
    \setlength{\tabcolsep}{2.5pt}
    \renewcommand{\arraystretch}{1.05}
    \caption{Quantitative comparison on \textbf{OVOW-4D-Scene-Bench}.}
    \label{tab:quant_4d}
    \begin{tabularx}{\linewidth}{@{}l*{7}{Y}@{}}
        \toprule
        Method & \hd{Scene-IoU}{AABB\,$\uparrow$} & \hd{Scene-IoU}{OBB\,$\uparrow$} & \hd{Object}{IoU\,$\uparrow$} & \hd{PL}{$\downarrow$} & \hd{N-CLIP}{$\downarrow$} & \hd{Time}{{\tiny(s/frame)}\,$\downarrow$} & \hd{VRAM}{{\tiny(GB)}\,$\downarrow$} \\
        \midrule
        CAST~\cite{cast}              & 0.091 & 0.176 & 0.080 & 13.10 & 1.74 & 365 & \cellcolor{cellbest}\textbf{10.5} \\
        SAM3D~\cite{sam3d}            & 0.055 & 0.135 & 0.050 & 10.20 & 1.98 & \cellcolor{cellsecond}103 & 33.5 \\
        VIGA~\cite{viga}              & \cellcolor{cellsecond}0.127 & 0.172 & 0.016 & 24.10 & 2.49 & 788 & 31.5 \\
        MIDI~\cite{midi}              & 0.095 & 0.198 & \cellcolor{cellsecond}0.174 & \cellcolor{cellsecond}5.60 & 1.56 & 234 & 17.7 \\
        SceneGen~\cite{mengscenegen}  & 0.096 & 0.183 & 0.051 & 8.20 & \cellcolor{cellsecond}1.54 & 152 & 29.6 \\
        TabletopGen~\cite{tabletopgen}& 0.072 & \cellcolor{cellsecond}0.200 & 0.034 & 7.50 & 1.91 & 638 & \cellcolor{cellsecond}14.0 \\
        \midrule
        \textbf{OVOW (Ours)}          & \cellcolor{cellbest}\textbf{0.180} & \cellcolor{cellbest}\textbf{0.440} & \cellcolor{cellbest}\textbf{0.210} & \cellcolor{cellbest}\textbf{2.90} & \cellcolor{cellbest}\textbf{1.43} & \cellcolor{cellbest}\textbf{3.35} & 26.0 \\
        \bottomrule
    \end{tabularx}
\end{table}

\subsection{Quantitative Comparison}
\label{sec:quant}

Tabs.~\ref{tab:quant_static} and~\ref{tab:quant_4d} report the comparison on the two benchmarks; in both, the \colorbox{cellbest}{best} and \colorbox{cellsecond}{second-best} per column are highlighted.
On OVOW-3D-Scene-Bench, \methodname obtains the best Scene-IoU-OBB (0.218), Object-IoU (0.190), PL (5.70), and N-CLIP (1.87); VIGA attains a higher AABB-IoU but trails on every other metric, and OVOW's single-image runtime (272\,s) is comparable to the feed-forward baselines.
On OVOW-4D-Scene-Bench, \methodname leads every quality metric, reaching 0.440 Scene-IoU-OBB, 0.210 Object-IoU, 2.90 PL, and 1.43 N-CLIP, while amortizing computation across the video to run at 3.35\,s per frame, one to two orders of magnitude faster than the baselines (103--788\,s).

\noindent\textbf{Where the gains come from.}
\methodname leads on both benchmarks, with the largest margin on the dynamic 4D scenes, where video-level temporal cues are most informative. Amodal inpainting keeps per-object geometry faithful under occlusion, and physics-grounded assembly resolves ground contact and inter-object support, so scenes show far fewer interpenetration and floating artifacts than the generative baselines and stay stable under gravity in a physics simulator.

\noindent\textbf{Scalability.}
Performance degrades gracefully as the number of objects per scene grows; the mild quality drop on the most crowded scenes is primarily caused by increased mutual occlusion among instances, while per-frame runtime grows only modestly.

\subsection{Qualitative Comparison}
\label{sec:qual}

\begin{figure*}[htbp]
    \centering
    \includegraphics[width=\linewidth]{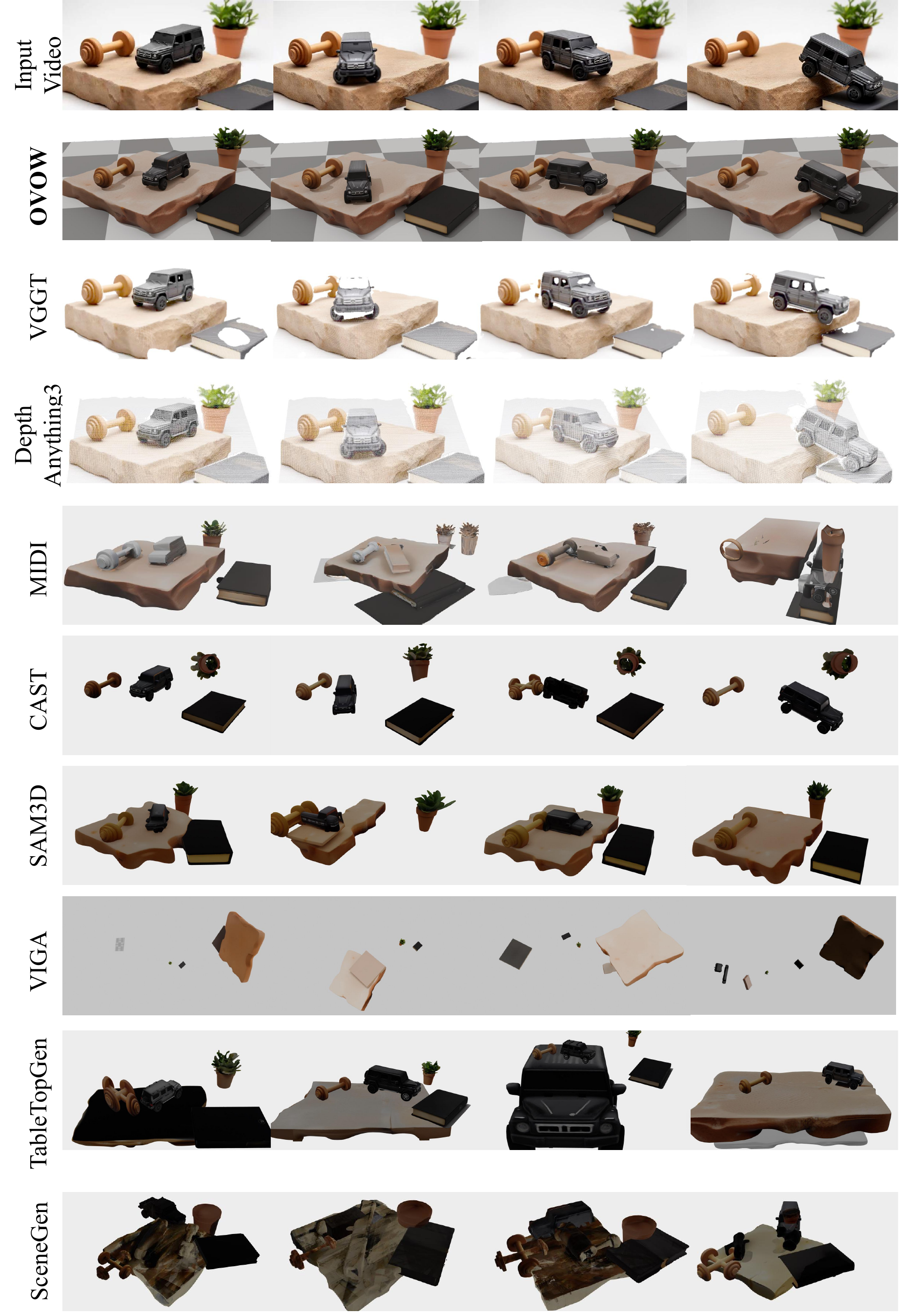}
    \caption{\textbf{Qualitative comparison on an in-the-wild dynamic scene.} Four frames of a dynamic scene comparing \methodname against feed-forward 3D reconstruction (VGGT~\cite{vggt}, Depth Anything~3~\cite{depthv3}) and instance-level scene-reconstruction methods.}
    \label{fig:comp_4d}
\end{figure*}

\begin{figure*}[htbp]
    \centering
    \includegraphics[width=\linewidth]{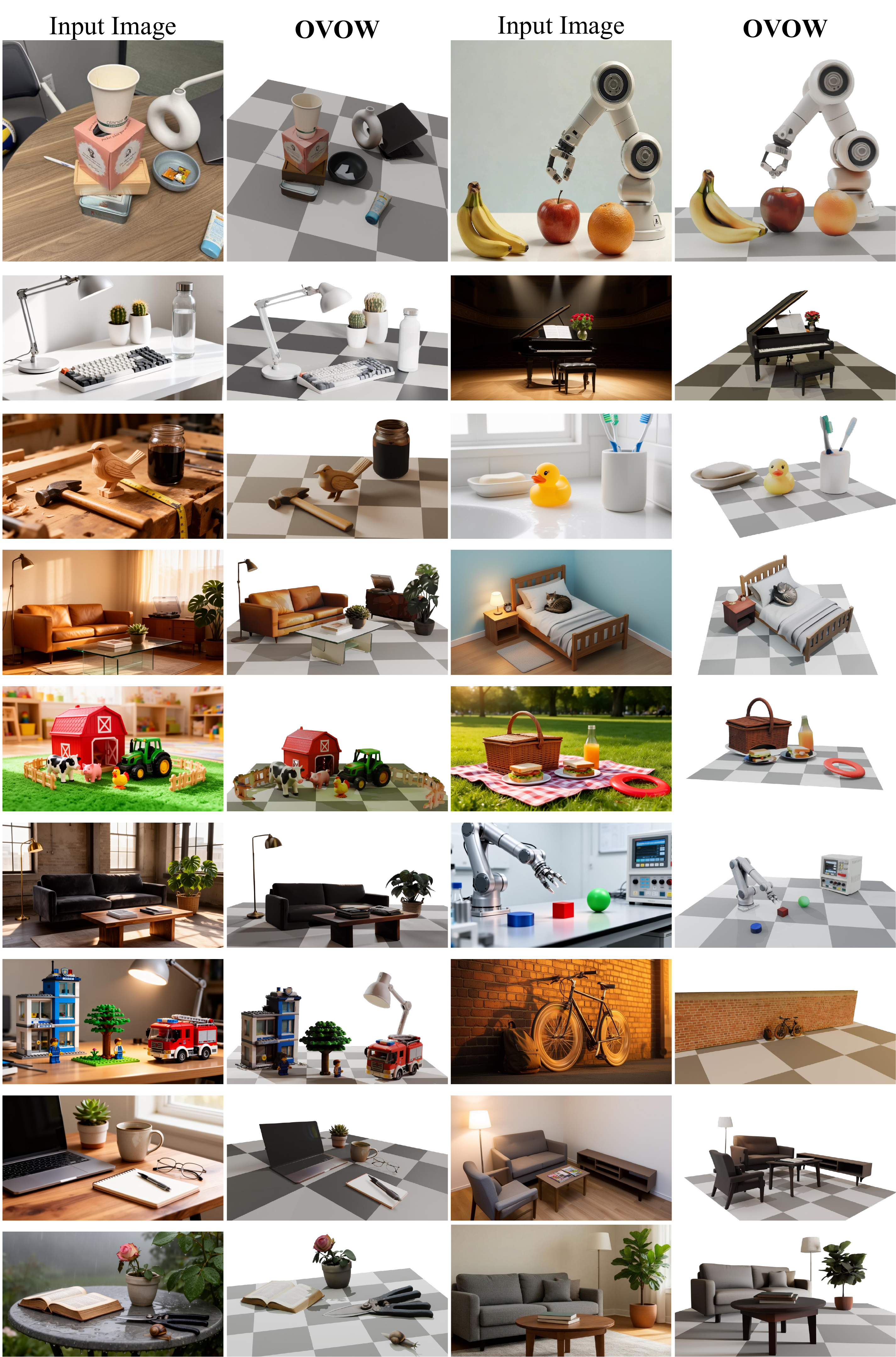}
    \caption{\textbf{Image-to-3D reconstruction by \methodname.}}
    \label{fig:demo_3d}
\end{figure*}

\begin{figure*}[htbp]
    \centering
    \includegraphics[width=\linewidth]{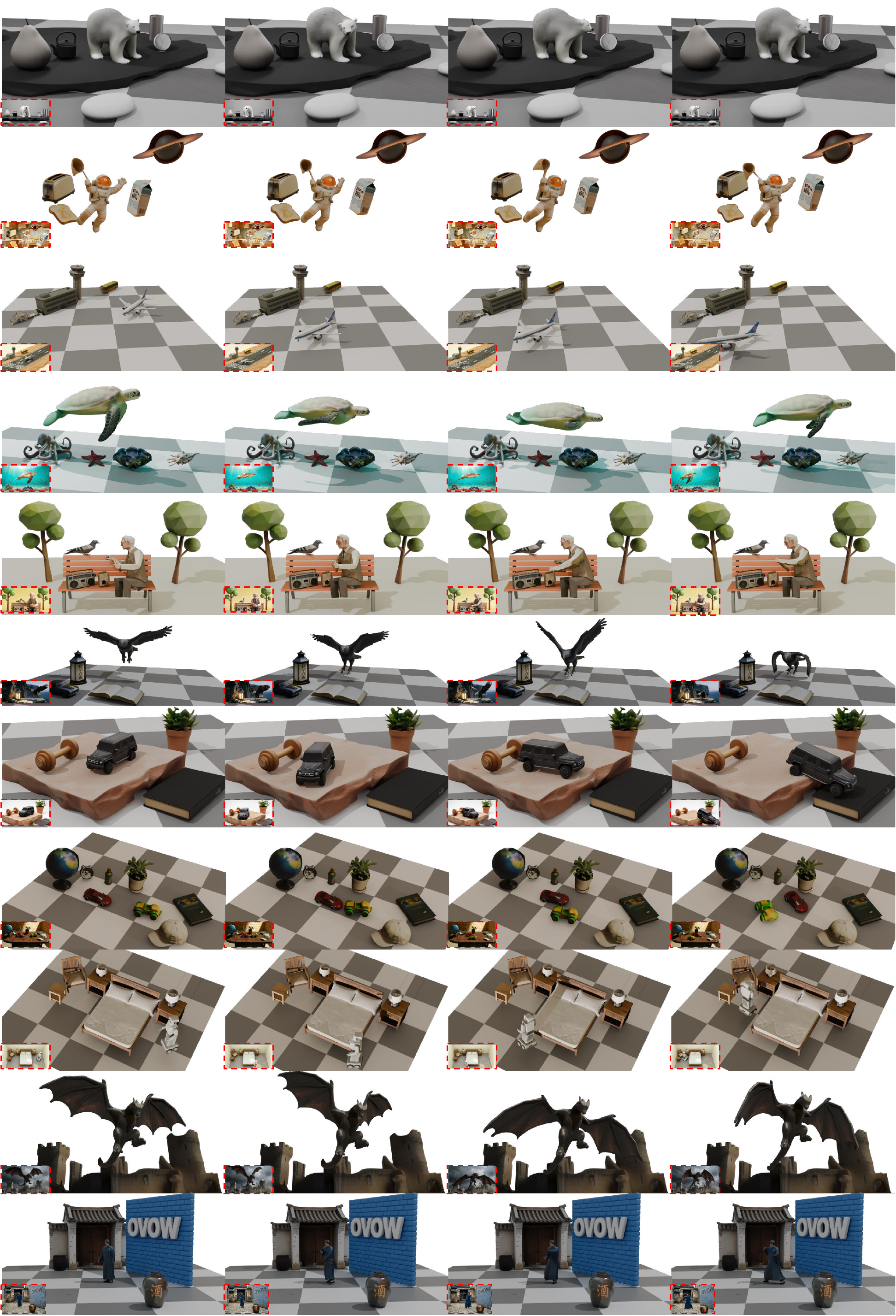}
    \caption{\textbf{Paired video-to-4D results generated by \methodname.} Each example is a reconstructed instance-level 4D mesh scene; the red box at the bottom-left corner of each render shows the corresponding input frame. 
    }
    \label{fig:dataset}
\end{figure*}

\begin{figure}[htbp]
    \centering
    \includegraphics[width=\linewidth]{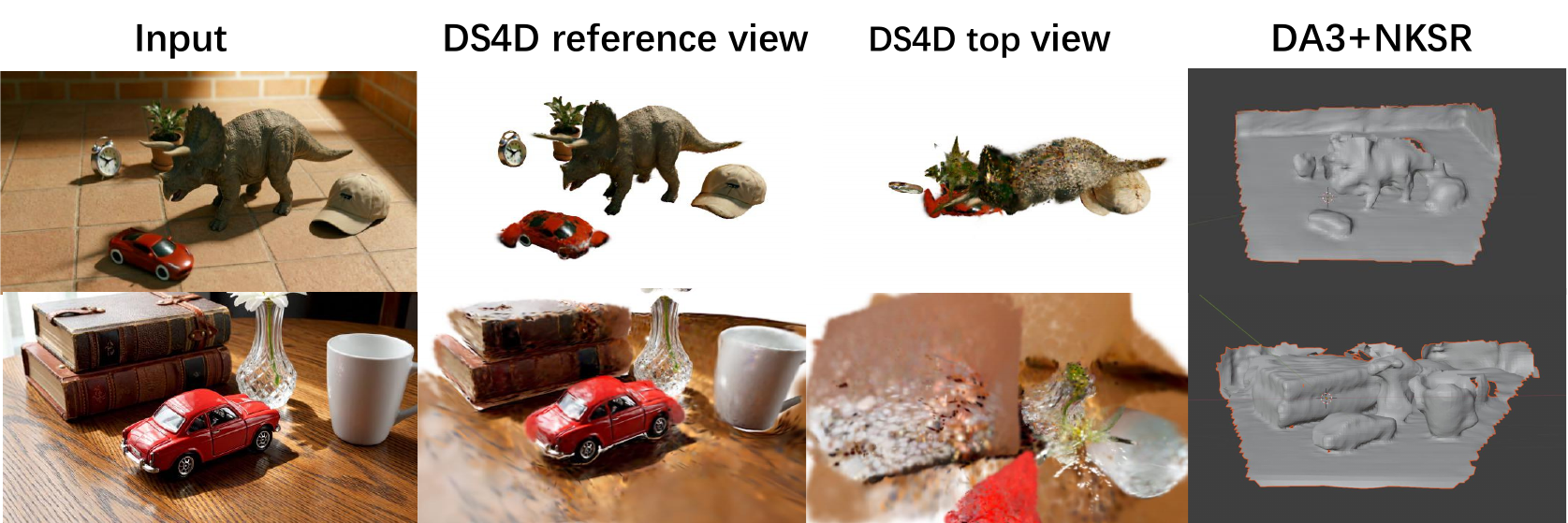}
    \caption{\textbf{Comparison with video-to-mesh and fused-surface alternatives.} DreamScene4D~\cite{dreamscene4d} (DS4D) is plausible at the reference view but degrades at other viewpoints, while Depth Anything~3~\cite{depthv3} followed by NKSR~\cite{nksr} (DA3+NKSR) fuses a single surface without instance separation, metric scale, or contact. \methodname instead outputs instance-separated, simulation-ready meshes.}
    \label{fig:comp_alt}
\end{figure}

Fig.~\ref{fig:comp_4d} compares \methodname against feed-forward 3D reconstruction (VGGT, Depth Anything~3) and instance-level scene-reconstruction baselines on a dynamic scene: \methodname uniquely recovers instance-separated, watertight meshes with accurate layout, while the feed-forward methods lack instance separation and the scene-reconstruction baselines misplace objects or distort geometry.
Fig.~\ref{fig:demo_3d} shows \methodname applied to single images (image-to-3D): from one in-the-wild frame it recovers instance-separated meshes with faithful shapes and accurate layout through iterative scale-orientation refinement, with per-baseline comparisons deferred to the appendix.
\methodname is the only method that simultaneously achieves accurate geometry, correct spatial layout, and watertight, simulation-ready topology, benefiting from video-level temporal consistency for scale and pose recovery and from amodal inpainting for occluded instances.
Fig.~\ref{fig:comp_alt} further contrasts \methodname with two tempting alternatives: single-object video-to-mesh generation (DreamScene4D) degrades at unseen viewpoints and recovers neither metric scale nor inter-object contact, while depth-fusion (Depth Anything~3 followed by NKSR) yields a single fused surface without instance separation.
Beyond per-scene reconstruction, Fig.~\ref{fig:dataset} shows paired video-to-4D examples that \methodname generates across diverse scene types, illustrating the data our pipeline can produce at scale.
Additional per-scene visualizations and comparisons are provided in the appendix.

\subsection{Sensitivity of Key Design Choices}
\label{sec:ablation}

\noindent\textbf{Hyperparameter robustness.}
On a held-out validation set spanning static, rigid, and deformable motion, \methodname is non-brittle to its key design choices (Tab.~\ref{tab:hparam}). Refinement iterations are the accuracy--runtime elbow (IoU-B saturates at $0.78$ by $N_{\text{iter}}{=}3$); $\rho{=}0.65$ and $\tau_{\text{area}}{=}200$\,px are near-optimal; and the contact parameters ($\epsilon{=}0.2\%$, $\delta_{\max}{=}2\%$, $p_{\text{bot}}{=}10\%$) share $\tau_{\text{contact}}$'s flat stability plateau.

\begin{table}[htbp]
    \centering
    \footnotesize
    \setlength{\tabcolsep}{4pt}
    \renewcommand{\arraystretch}{1.05}
    \caption{\textbf{Hyperparameter robustness} on the validation set (spanning static, rigid, and deformable motion). Defaults are \underline{underlined}; ``\,--\,'' marks unused cells. Metrics are bounding-box IoU (IoU-B), valid-scene rate (\%), and simulation stability (\%).}
    \label{tab:hparam}
    \begin{tabular}{llcccc}
        \toprule
        Hyperparameter & Metric & \multicolumn{4}{c}{Values (default \underline{underlined})} \\
        \midrule
        $N_{\text{iter}}\!\in\!\{1,2,\underline{3},5\}$        & IoU-B$\uparrow$           & 0.58 & 0.72 & \textbf{0.78} & 0.79 \\
        $\rho\!\in\!\{0.55,\underline{0.65},0.75\}$            & IoU-B$\uparrow$           & 0.74 & \textbf{0.78} & 0.76 & --   \\
        $\tau_{\text{area}}\!\in\!\{100,\underline{200},400\}$ & Valid scenes$\uparrow$    & 84.9 & \textbf{86.8} & 85.7 & --   \\
        $\tau_{\text{contact}}\!\in\!\{2,\underline{3},4\}\%$  & Sim.\ stability$\uparrow$ & 79.8 & \textbf{82.7} & 81.5 & --   \\
        $N_{\text{asm}}\!\in\!\{1,\underline{2},5\}$            & Sim.\ stability$\uparrow$ & 78.9 & \textbf{82.7} & 82.8 & --   \\
        \bottomrule
    \end{tabular}
\end{table}

\noindent\textbf{Stage-level reliability.}
The pipeline stages are also reliable on this validation set: $95.4\%$ motion-category accuracy, $93.1\%/88.7\%$ rigid/deformable reconstruction success, $92.4\%$ pose recovery, and $86.8\%$ final validity with $82.7\%$ simulation stability, so failures rarely propagate.

\section{Conclusion}
\label{sec:conclusion}

We presented \methodname, the first training-free system that turns monocular video into instance-level, simulation-ready 4D mesh scenes, modeling all motion via direct vertex deformation without rigging or category priors.
It leads the baselines in geometry, layout, and photometric/semantic accuracy, runs one to two orders of magnitude faster on video, and doubles as a scalable engine for paired video-to-4D simulation data.

\bibliographystyle{splncs04}
\bibliography{main}

\clearpage
\appendix

\setcounter{section}{0}
\renewcommand{\thesection}{\Alph{section}}
\renewcommand{\thetable}{\Alph{section}\arabic{table}}
\renewcommand{\thefigure}{\Alph{section}\arabic{figure}}

\section{Details of the Evaluation Benchmarks}
\setcounter{table}{0}
\setcounter{figure}{0}
\label{sec:supp_benchmark}

\noindent\textbf{Motivation for Synthetic Evaluation Data.}
A central challenge in evaluating structured Video-to-4D reconstruction is the absence of real-world datasets that pair source videos with ground-truth instance-level 4D mesh annotations.
Existing video benchmarks~\cite{co3d,objaverse} either lack instance-level mesh decomposition or provide only rendering-oriented representations (e.g., point clouds, depth maps).
To enable rigorous quantitative evaluation of geometric correctness, instance separation quality, and physical plausibility, we construct a fully synthetic benchmark with precise ground-truth control over all aspects of the scene.

\begin{figure}[htbp]
    \centering
    \includegraphics[width=1\linewidth]{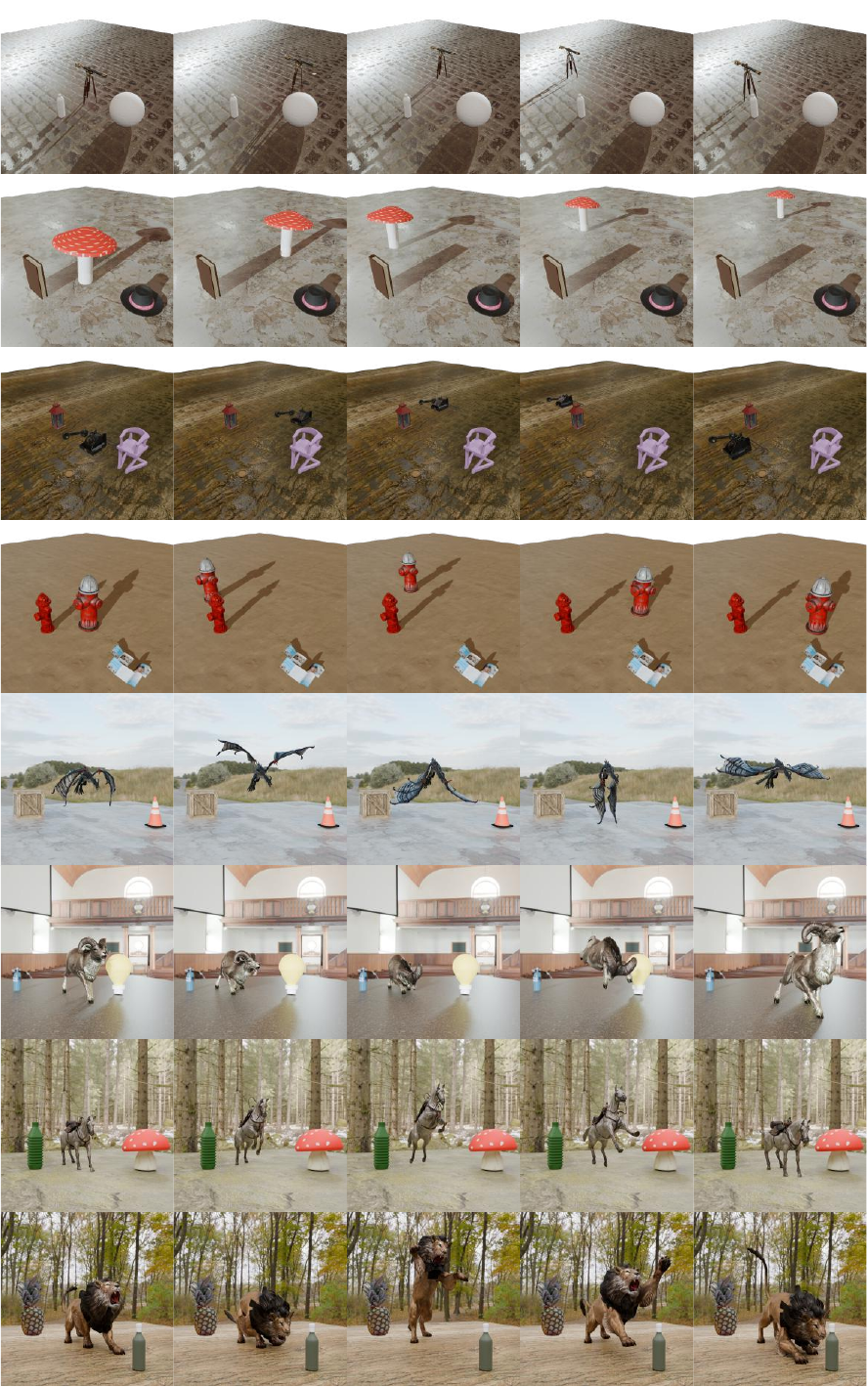}
    \caption{\textbf{Representative scenes our synthetic 3D/4D scene-data pipeline can generate.} The pipeline composes existing 3D meshes, animated assets, and HDRI environments into scenes with diverse object categories, layouts, lighting, and motion types (static, rigid, deformable). \emph{These illustrate the pipeline's range; the quantitative benchmark draws a controlled static\,+\,rigid subset from it.} Each scene provides per-instance segmentation masks, watertight meshes with 6-DoF pose trajectories, and dense depth maps.}
    \label{fig:eval_data_demo}
\end{figure}

\noindent\textbf{The synthetic scene-data pipeline.}
Distinct from \methodname, which reconstructs 4D scenes from \emph{real} video, we build a separate pipeline that \emph{composes} 3D/4D scenes from existing assets in Blender, yielding exact ground truth for quantitative evaluation. It proceeds in four steps:

\begin{enumerate}[leftmargin=1.5em,itemsep=2pt]
    \item \textbf{Asset Sampling.} We draw 3D object meshes from Objaverse-OA~\cite{lu2025orientation} (orientation-aligned, spanning furniture, vehicles, tableware, toys, etc.), animated 4D assets from the Truebones Zoo dataset\footnote{\url{https://truebones.gumroad.com/}} (motion-captured animals), and HDRI environments from Poly Haven\footnote{\url{https://github.com/Poly-Haven/polyhavenassets}}.

    \item \textbf{Scene Composition.} For each scene we randomly place \textbf{3--5} object instances on a ground plane with randomized, physically plausible, non-overlapping positions and orientations, under a sampled HDRI environment.

    \item \textbf{Motion Animation.} The pipeline can render static, rigid-body, and deformable motion (Fig.~\ref{fig:eval_data_demo}). \textbf{For the evaluation benchmark we use static and rigid-body motion only}, for which per-frame ground-truth poses are exact: every OVOW-4D-Scene-Bench scene contains at least one rigid mover (translation and rotation keyframed along procedural linear, circular, and random-walk trajectories) while the rest stay static, whereas OVOW-3D-Scene-Bench scenes contain no motion.

    \item \textbf{Rendering and Annotation.} Each scene is rendered as a 108-frame video at $512 \times 512$, exporting per-frame ground truth simultaneously: (a) instance-level segmentation masks, (b) per-object watertight meshes with 6-DoF pose trajectories, (c) camera intrinsics and extrinsics, and (d) dense depth maps. This enables all metrics reported in the main paper.
\end{enumerate}

\noindent\textbf{Benchmark subsets.}
The benchmark comprises \emph{OVOW-3D-Scene-Bench}, with 120 fully static scenes, and \emph{OVOW-4D-Scene-Bench}, with 120 dynamic scenes that each contain at least one rigidly-moving object alongside static ones, every scene comprising a random 3--5 instances. We restrict the \emph{quantitative test} benchmark to static and rigid motion, for which ground-truth meshes and trajectories are exact; deformable motion is additionally exercised on a held-out synthetic \emph{validation} set (used for the ablation and hyperparameter studies) and assessed qualitatively on real-world videos (Sec.~\ref{sec:supp_quant}).

\noindent\textbf{Quality Assurance.}
We manually inspect every evaluation scene to verify: (1) all object instances are correctly labeled with appropriate motion categories; (2) no visual artifacts exist in the rendered videos (e.g., interpenetration, floating objects, or lighting inconsistencies); (3) the ground-truth annotations (masks, poses, meshes) are accurate and temporally consistent. Scenes that do not pass inspection are discarded and regenerated.

\noindent\textbf{Synthetic-to-Real Generalizability.}
While our quantitative evaluation is conducted on synthetic data, we emphasize that our method operates in a fully training-free manner, composing pre-trained foundation models without any fine-tuning on our benchmark. As demonstrated by the qualitative results in the main paper (Figs.~\ref{fig:comp_4d} and~\ref{fig:demo_3d}) and the additional comparisons in Section~\ref{sec:supp_quant}, \methodname generalizes well to diverse real-world inputs. The synthetic benchmark serves as a controlled testbed for precise quantitative measurement, while real-world results validate practical applicability. Fig.~\ref{fig:eval_data_demo} showcases the range of scenes our synthetic pipeline can produce.

\FloatBarrier
\section{Details of the OVOW Video-to-4D Dataset}
\setcounter{table}{0}
\setcounter{figure}{0}
\label{sec:supp_dataset}

A key contribution of our work is a scalable pipeline for generating paired videos and their corresponding instance-level 4D mesh scenes. Because the pipeline is fully automated and training-free, it imposes no hard cap and can synthesize high-quality video--4D scene pairs at scale. This section describes the data curation pipeline in detail.

\noindent\textbf{Data Curation Pipeline.}
Our dataset construction follows a six-stage pipeline:

\begin{enumerate}[leftmargin=1.5em,itemsep=2pt]
    \item \textbf{Prompt Generation.}
    We design a structured text prompt template that parameterizes scene composition along multiple axes: (a) a curated vocabulary of rigid objects (furniture, vehicles, tools, tableware, etc.) and deformable objects (animals, cloth, plants, humans, etc.); (b) the number of objects per scene, uniformly sampled from 3 to 10; (c) lighting conditions (studio, natural daylight, sunset, overcast, etc.); (d) environment types (indoor, outdoor, tabletop, urban, nature, etc.); and (e) camera viewpoints (eye-level, top-down, low-angle, etc.). We generate random combinations of these variables and use Qwen3-VL~\cite{qwen3vl} to produce descriptive text prompts suitable for image generation.

    \item \textbf{Image Generation.}
    We use FLUX.2~\cite{flux2} for text-to-image generation, producing high-resolution ($1024 \times 1024$) images from the generated prompts. Each image depicts a multi-object scene with the specified composition.

    \item \textbf{Image-Level Filtering.}
    We apply Qwen3-VL as a visual quality filter, discarding images that: (a) contain fewer than 3 identifiable object instances; (b) do not depict a coherent scene suitable for 3D reconstruction (e.g., abstract art, text-heavy images, close-up single objects); or (c) exhibit severe visual artifacts. This stage removes approximately 35\% of generated images.

    \item \textbf{Video Generation.}
    Filtered images are converted to videos using Wan2.2, which generates temporally coherent 108-frame videos with plausible object motions from the input images.

    \item \textbf{Video-Level Filtering.}
    We execute the VLM-Guided Scene Decomposition stage (Stage~1) of the \methodname pipeline on each generated video to verify that: (a) the video contains at least one moving object (rigid or deformable); and (b) the total number of segmented instances is $\geq 3$. Videos that do not meet these criteria are discarded.

    \item \textbf{4D Scene Reconstruction and Quality Control.}
    The full \methodname pipeline converts each filtered video into an instance-level 4D mesh scene. We then apply automated quality filters to remove scenes with anomalous photometric or semantic-alignment scores (exceeding 2 standard deviations from the mean), followed by manual screening to ensure geometric plausibility and visual quality.
\end{enumerate}

\noindent\textbf{Downstream Applications.}
The proposed Video-to-4D dataset can serve as training data for several downstream tasks: (1) learning-based 4D scene generation from video; (2) fine-grained scene understanding with instance-level mesh decomposition; (3) embodied AI data augmentation with simulation-ready assets; and (4) 4D world model pre-training.

\FloatBarrier
\section{Video to 4D Task}\label{sec:supp_quant}
\setcounter{table}{0}
\setcounter{figure}{0}

\subsection{Additional Qualitative Results}

\noindent\textbf{Comparison with Point-Cloud-Based Video Reconstruction Methods.}
A natural question is why we do not compare with point-cloud-based video-to-3D/4D methods such as VGGT~\cite{vggt} and Depth Anything~3~\cite{depthv3}. We clarify the distinction here and provide visual comparisons to illustrate why direct quantitative comparison is not appropriate.

Our method reconstructs \emph{instance-level watertight meshes} for individual objects in the scene, whereas point-cloud-based methods reconstruct \emph{dense scene-level point clouds} that include both foreground objects and background (walls, floors, sky, etc.). The key differences are:

\begin{enumerate}[leftmargin=1.5em,itemsep=2pt]
    \item \textbf{Representation mismatch.} Our evaluation metrics (Volumetric IoU, Chamfer Distance, F-Score) are computed on per-instance meshes using voxelized representations. Point clouds lack closed surface topology and cannot be directly evaluated with these metrics without non-trivial mesh reconstruction steps (e.g., Poisson surface reconstruction), which would introduce additional errors unrelated to the reconstruction method itself.

    \item \textbf{Scope mismatch.} Point-cloud methods reconstruct the entire visible scene including backgrounds, while \methodname reconstructs only the foreground object instances. Pixel-level metrics (PSNR/SSIM) would naturally favor whole-scene methods due to their coverage of background regions, which our method does not attempt to reconstruct.

    \item \textbf{Task mismatch.} Point-cloud representations are not simulation-ready: they lack watertight topology, instance-level separation, and URDF-compatible interfaces. Our method targets a fundamentally different output format designed for downstream physics simulation and embodied AI applications.
\end{enumerate}

\begin{figure}[htbp]
    \centering
    \includegraphics[width=1\linewidth]{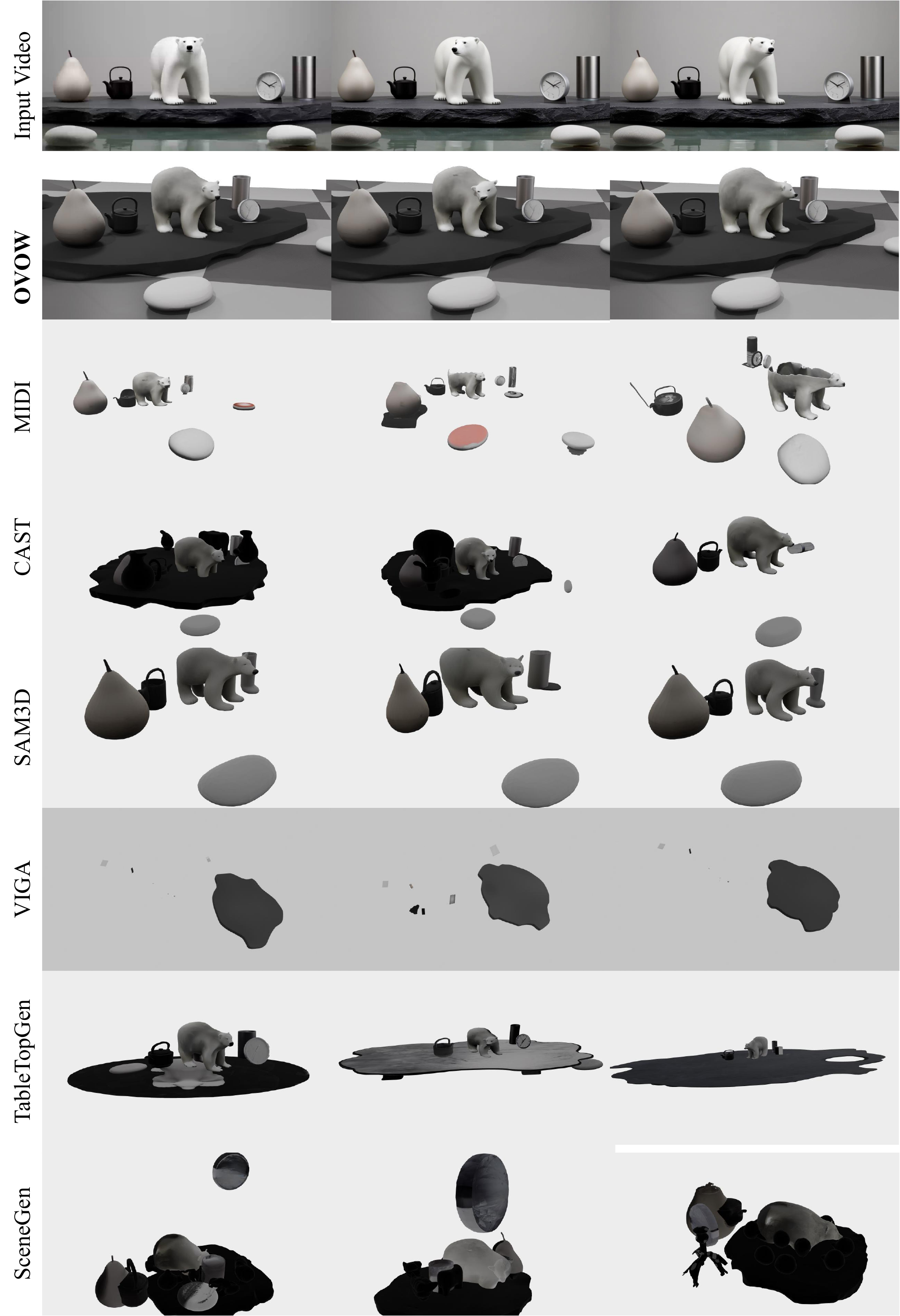}
    \caption{\textbf{Visual comparison with point-cloud-based methods (Case 1: polar bear).} The deformable polar bear is reconstructed as a topology-consistent mesh sequence by \methodname, capturing head-turning motion through per-vertex displacement fields. Point-cloud methods produce noisy, temporally inconsistent point clouds for this deformable object.}
    \label{fig:supp_comp_v4d_2}
\end{figure}

\begin{figure}[htbp]
    \centering
    \includegraphics[width=1\linewidth]{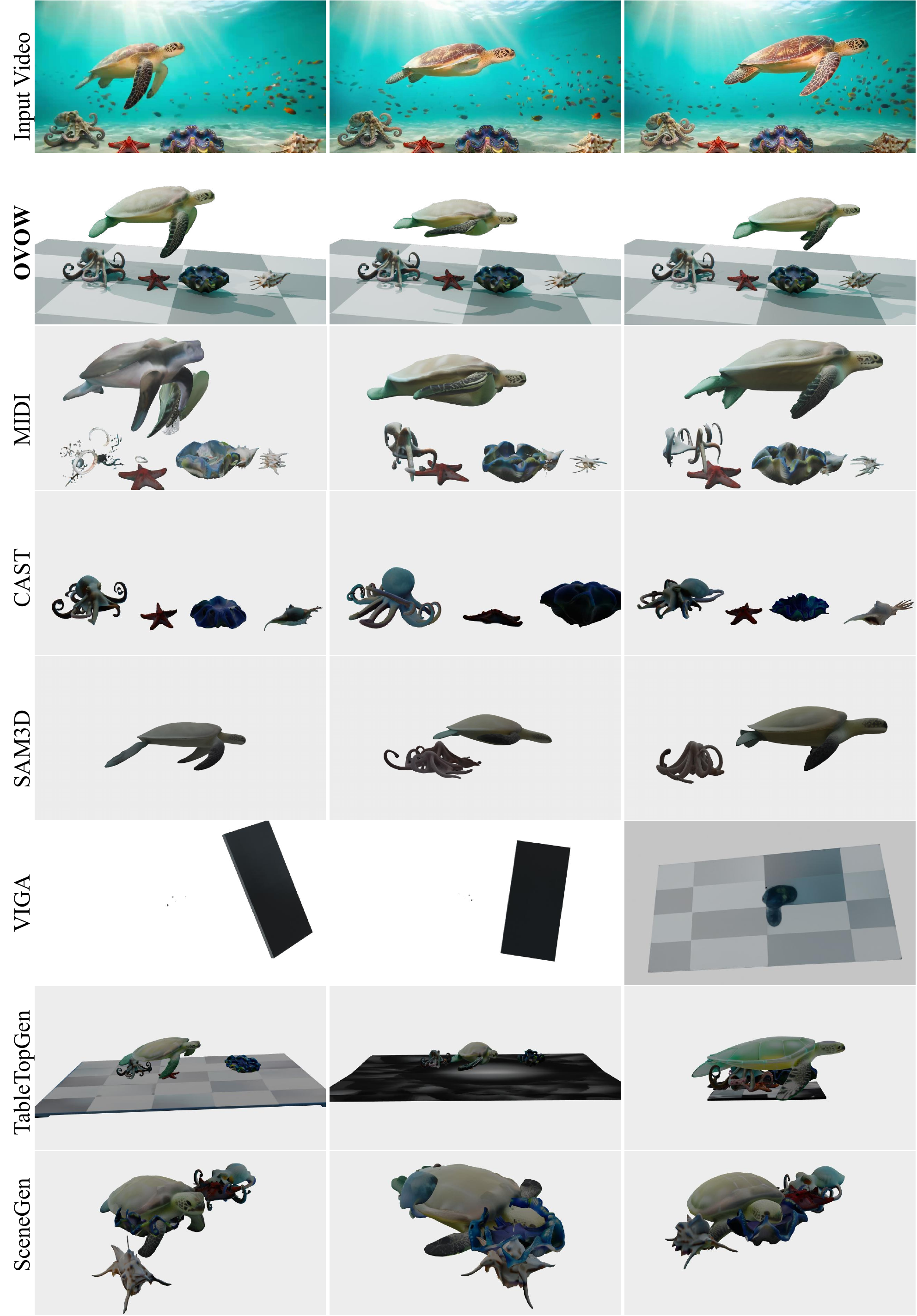}
    \caption{\textbf{Visual comparison with point-cloud-based methods (Case 2: sea turtle).} \methodname recovers the deformable sea turtle as a temporally coherent mesh sequence with consistent topology, while point-cloud methods produce scattered, unstructured point distributions that fail to preserve the object's surface continuity.}
    \label{fig:supp_comp_v4d_3}
\end{figure}

\begin{figure}[htbp]
    \centering
    \includegraphics[width=1\linewidth]{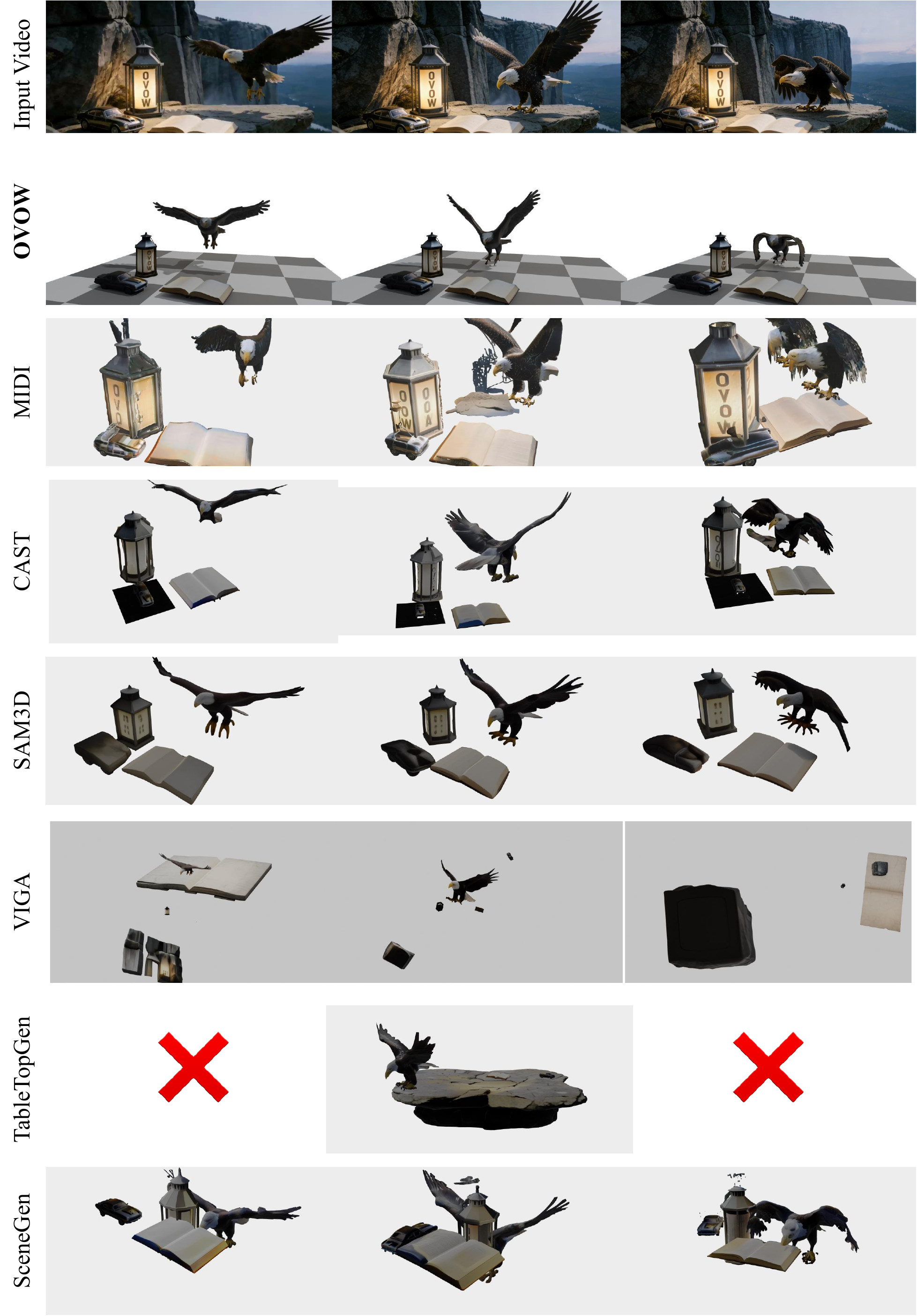}
    \caption{\textbf{Visual comparison with point-cloud-based methods (Case 3: eagle).} The flying eagle presents a challenging case with significant pose variation and deformation. \methodname captures the wing motion as a mesh sequence, whereas point-cloud methods produce sparse and noisy reconstructions that lose fine geometric details.}
    \label{fig:supp_comp_v4d_4}
\end{figure}

\noindent The black-car case in the main paper (Fig.~\ref{fig:comp_4d}) compares \methodname against two representative point-cloud-based methods, VGGT~\cite{vggt} and Depth Anything~3~\cite{depthv3}: point-cloud methods produce dense but unstructured reconstructions, whereas \methodname produces clean, instance-separated meshes with accurate geometry and texture. Figs.~\ref{fig:supp_comp_v4d_2}--\ref{fig:supp_comp_v4d_4} provide additional \methodname reconstructions on more in-the-wild videos.

\noindent\textbf{Comparison with Video-to-Mesh and Fused-Surface Alternatives.}
We also contrast \methodname with two other families of approaches that might appear applicable: single-object video-to-mesh generation (DreamScene4D~\cite{dreamscene4d}) and depth-fusion surface reconstruction (Depth Anything~3~\cite{depthv3} followed by NKSR~\cite{nksr}).
As shown in Fig.~\ref{fig:comp_alt} (main paper), DreamScene4D looks plausible from the reference view but exhibits unrealistic geometry from unseen viewpoints, and it recovers neither metric scale nor inter-object contact.
DA3+NKSR fuses a single continuous surface that lacks instance separation, watertight per-object topology, and physical grounding.
In contrast, \methodname produces instance-separated, simulation-ready meshes with recovered scale and contact, which these alternatives do not provide.

\FloatBarrier
\subsection{Additional Quantitative Results}

The main paper compares all seven methods on OVOW-3D-Scene-Bench and OVOW-4D-Scene-Bench (Tabs.~\ref{tab:quant_static} and~\ref{tab:quant_4d}). Here we summarize the qualitative trends behind those results for the baselines not discussed in detail in the main text.
Among the feed-forward baselines, MIDI~\cite{midi} and CAST~\cite{cast} produce reasonable per-object appearance but place objects less accurately within the scene, lowering their scene-level IoU.
TabletopGen~\cite{tabletopgen} benefits from a structured scene-generation prior for tabletop-style layouts but generalizes less well to the diverse scene types in our benchmark; its purely generative placements lack physics-grounded contact constraints, leading to more interpenetration.
SAM3D~\cite{sam3d}, built on a segmentation backbone, lacks explicit spatial-layout reasoning and is not designed for temporally coherent reconstruction, so it trails on the dynamic 4D scenes.

\noindent\textbf{Per-Category and Physical Plausibility.}
Across the benchmark, \methodname's advantage is largest on the dynamic 4D scenes, where video-level temporal information is most informative, and on physical plausibility. Because our physics-grounded assembly explicitly enforces ground contact and inter-object support, the reconstructed scenes show markedly less interpenetration and fewer floating objects than the generative baselines, and remain stable when dropped into a physics simulator under gravity.

\FloatBarrier
\section{Image to 3D Scene Task}
\setcounter{table}{0}
\setcounter{figure}{0}
\label{sec:supp_image23d}

While \methodname is primarily designed for video-to-4D reconstruction, it naturally supports image-to-3D scene reconstruction by treating a single image as a one-frame video. In this setting, the VLM-Guided Scene Decomposition identifies all object instances from the single image, and the pipeline proceeds with mesh reconstruction and physics-grounded assembly without the temporal pose tracking stage (all objects are treated as static).

\subsection{Additional Qualitative Results}

Beyond the image-to-3D demo results shown in the main paper (Fig.~\ref{fig:demo_3d}), we provide per-baseline qualitative comparisons below.

\begin{figure}[htbp]
    \centering
    \includegraphics[width=1\linewidth]{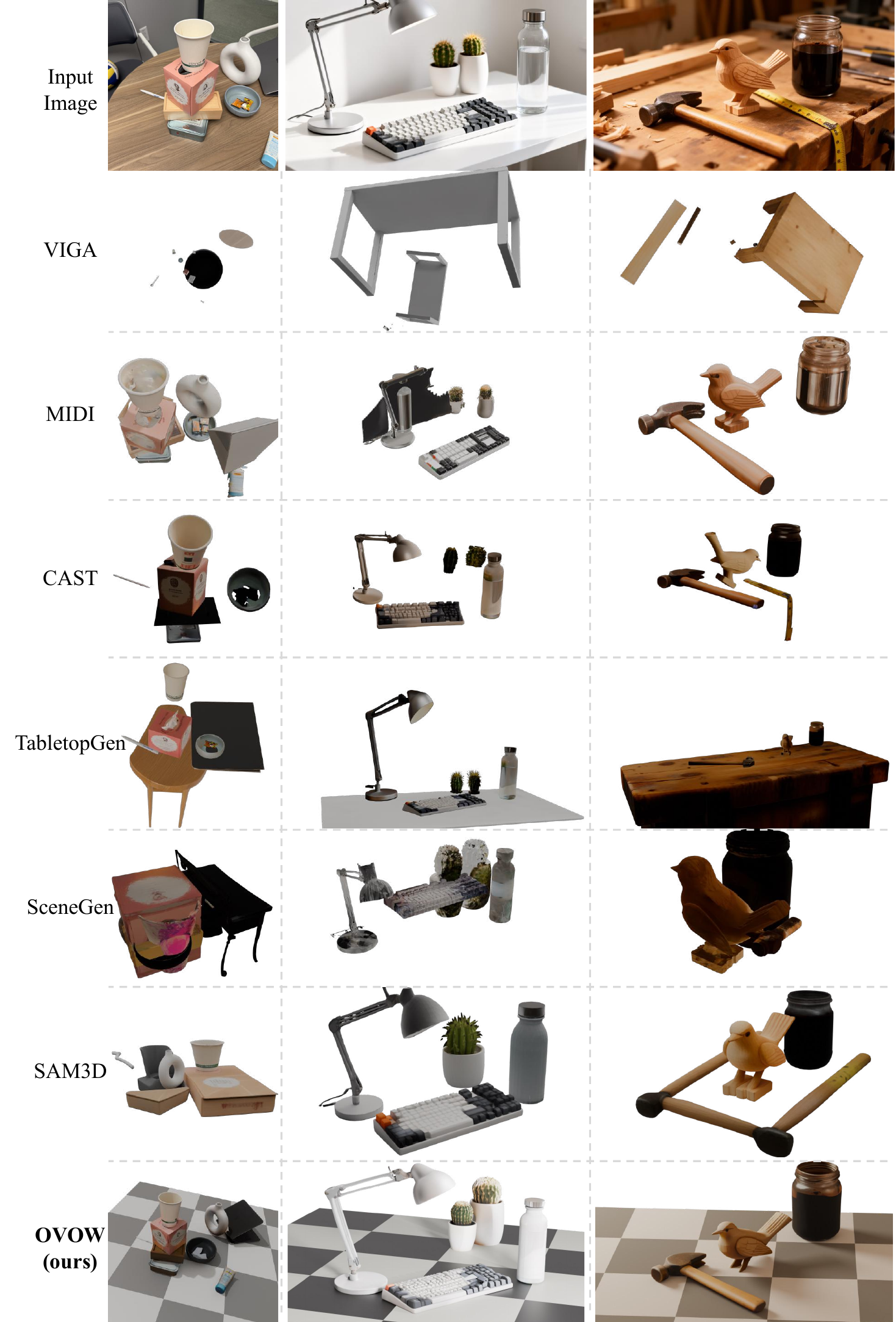}
    \caption{\textbf{Qualitative comparison on image-to-3D scene reconstruction (1/6).} We compare \methodname with VIGA, MIDI, CAST, TabletopGen, SceneGen, and SAM3D. Our method produces more complete object meshes with correct spatial arrangements.}
    \label{fig:supp_comp_i3d_1}
\end{figure}

\begin{figure}[htbp]
    \centering
    \includegraphics[width=1\linewidth]{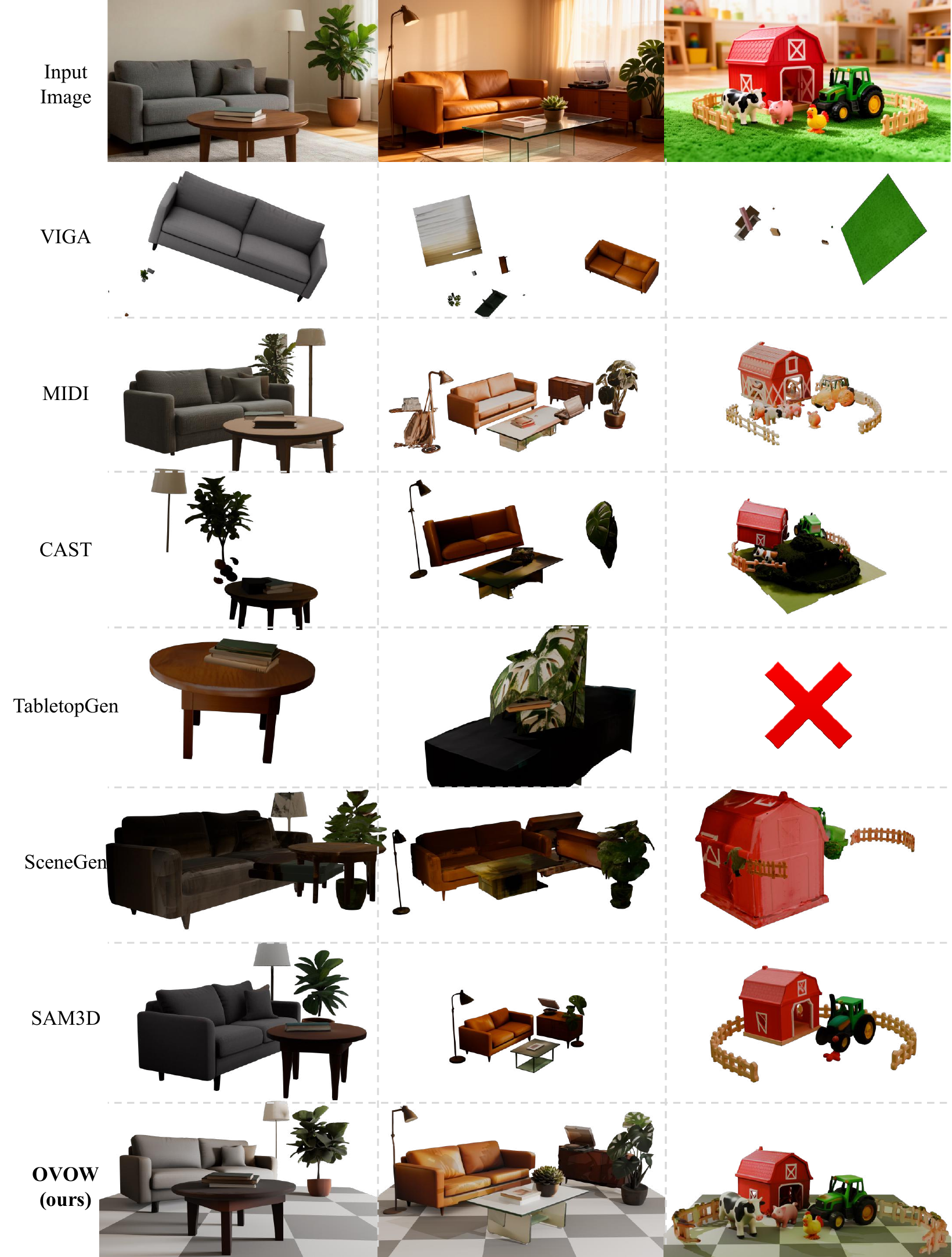}
    \caption{\textbf{Qualitative comparison on image-to-3D scene reconstruction (2/6).} \methodname benefits from amodal inpainting to handle partially occluded objects, producing more complete and accurate meshes than baselines.}
    \label{fig:supp_comp_i3d_2}
\end{figure}

\begin{figure}[htbp]
    \centering
    \includegraphics[width=1\linewidth]{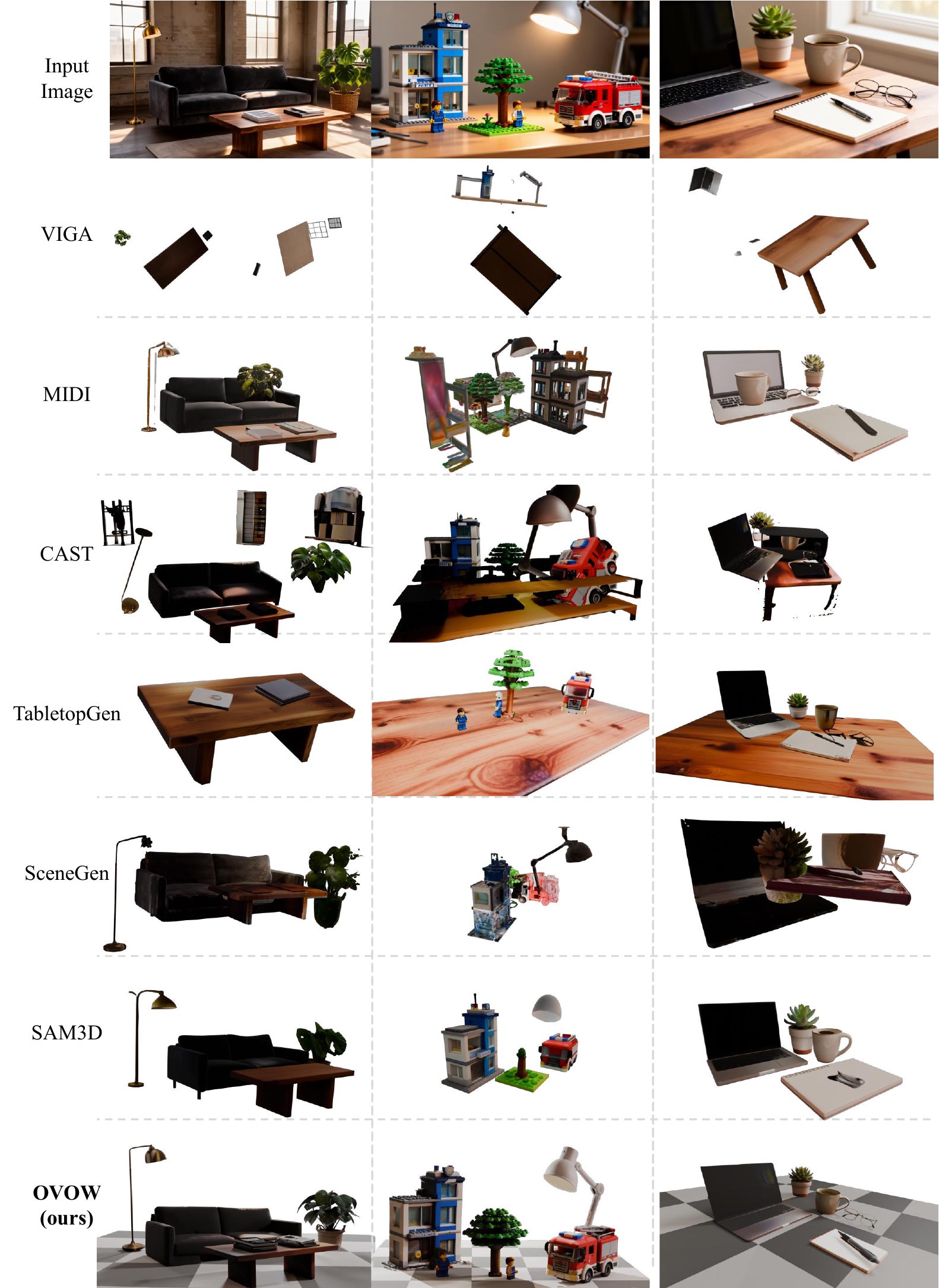}
    \caption{\textbf{Qualitative comparison on image-to-3D scene reconstruction (3/6).} In the single-image setting, baselines frequently produce distorted geometry or incorrect object scales, while \methodname recovers faithful shapes with correct spatial layout.}
    \label{fig:supp_comp_i3d_3}
\end{figure}

\begin{figure}[htbp]
    \centering
    \includegraphics[width=1\linewidth]{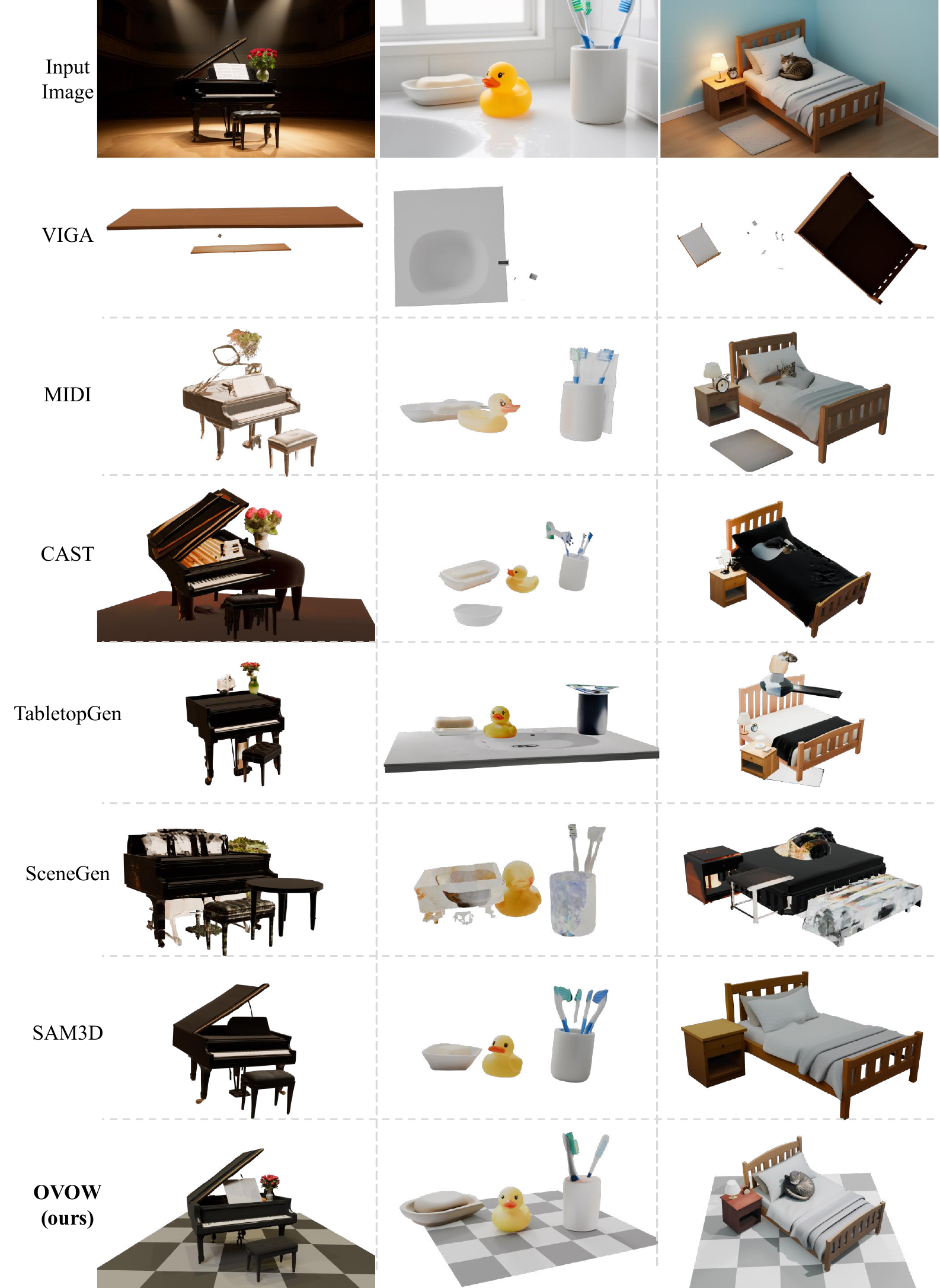}
    \caption{\textbf{Qualitative comparison on image-to-3D scene reconstruction (4/6).} For scenes with multiple overlapping objects, \methodname correctly separates instances and reconstructs each object with consistent geometry and texture.}
    \label{fig:supp_comp_i3d_4}
\end{figure}

\begin{figure}[htbp]
    \centering
    \includegraphics[width=1\linewidth]{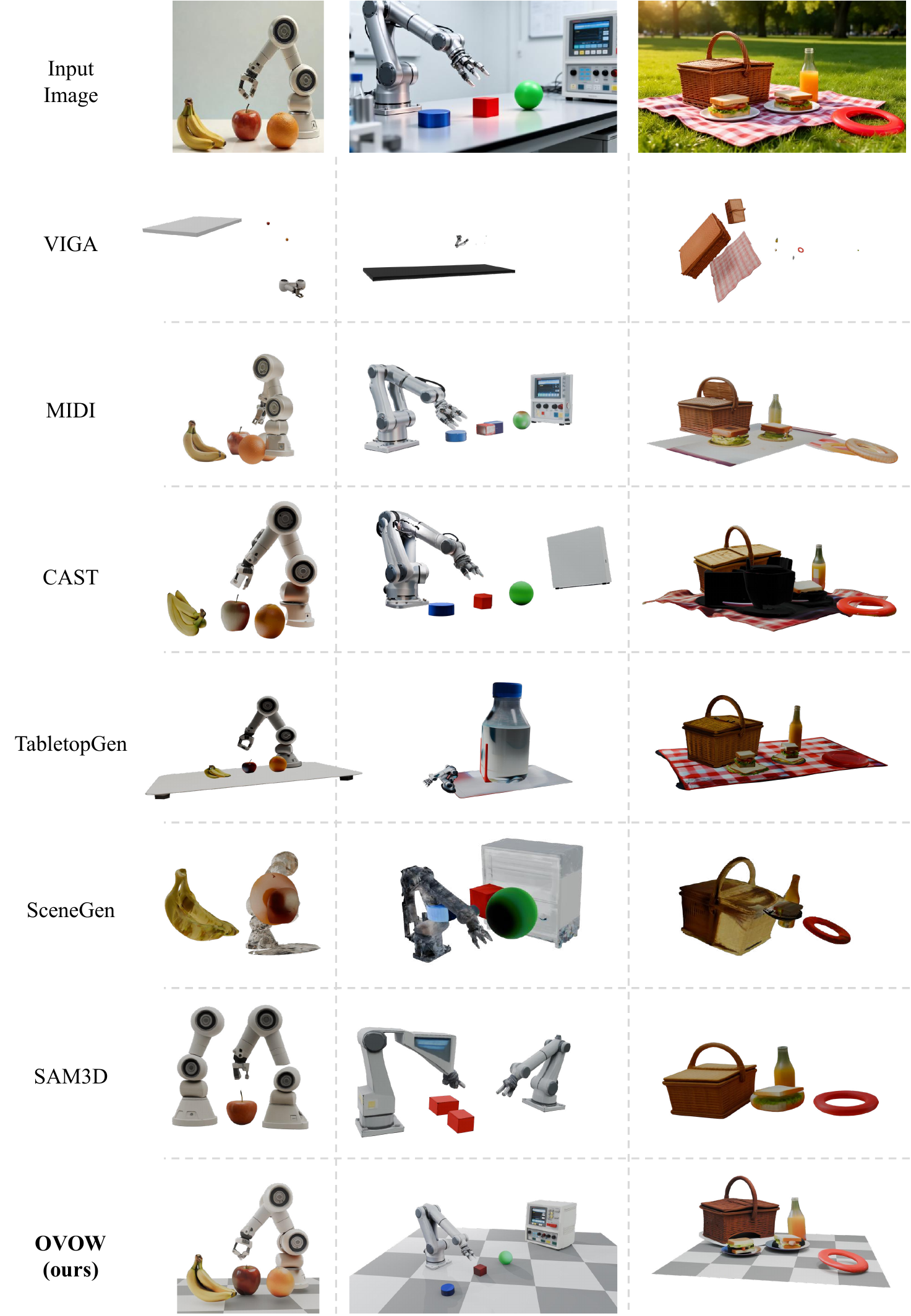}
    \caption{\textbf{Qualitative comparison on image-to-3D scene reconstruction (5/6).} \methodname handles diverse object categories including furniture, animals, and household items, producing high-quality meshes in all cases.}
    \label{fig:supp_comp_i3d_5}
\end{figure}

\begin{figure}[htbp]
    \centering
    \includegraphics[width=1\linewidth]{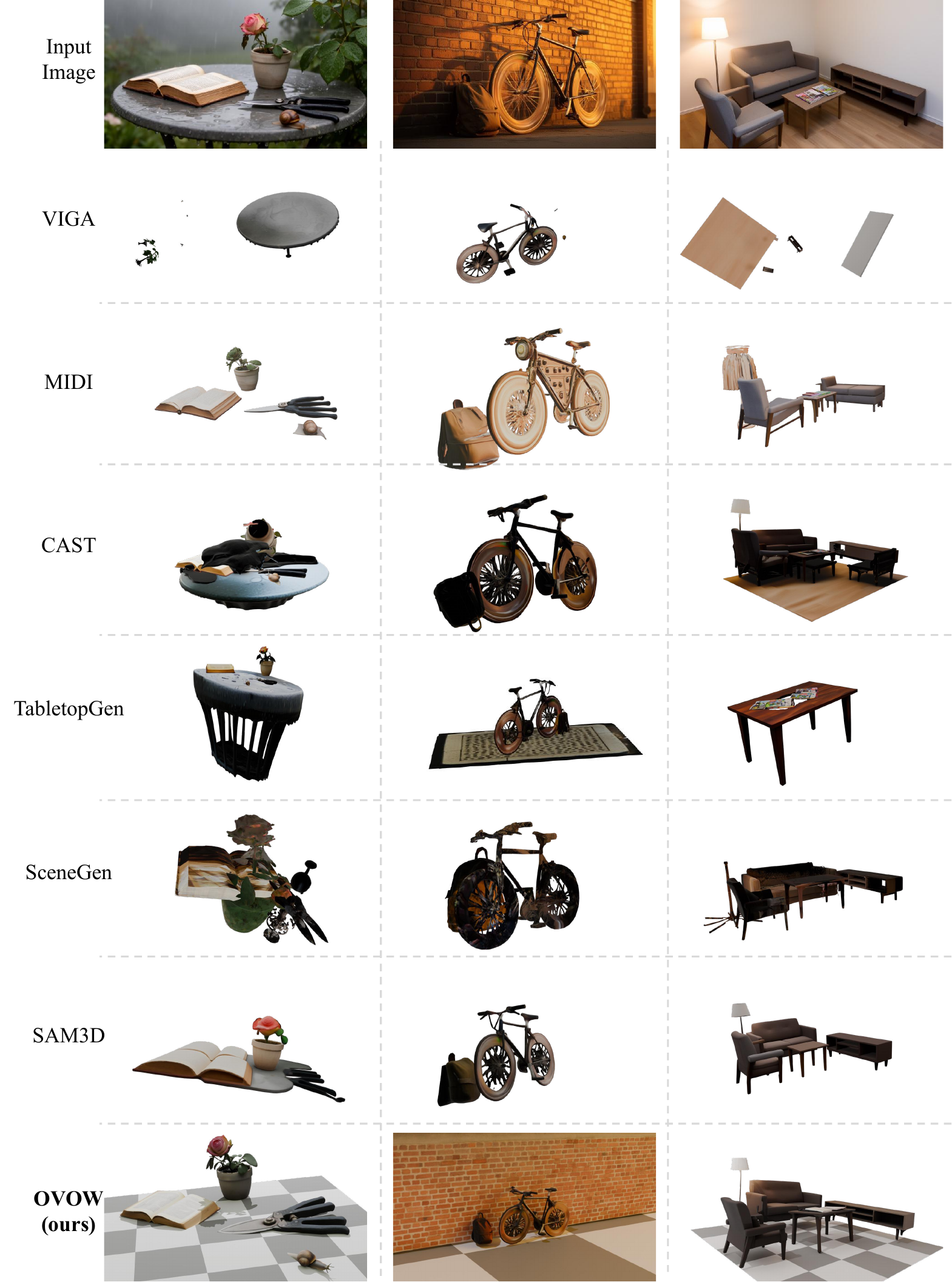}
    \caption{\textbf{Qualitative comparison on image-to-3D scene reconstruction (6/6).} Complex scenes with many objects and varied materials demonstrate the scalability and robustness of \methodname across diverse visual conditions.}
    \label{fig:supp_comp_i3d_6}
\end{figure}

Figs.~\ref{fig:supp_comp_i3d_1}--\ref{fig:supp_comp_i3d_6} present qualitative comparisons with six baselines (VIGA~\cite{viga}, MIDI~\cite{midi}, CAST~\cite{cast}, TabletopGen~\cite{tabletopgen}, SceneGen~\cite{mengscenegen}, and SAM3D~\cite{sam3d}) on the image-to-3D scene reconstruction task. \methodname consistently produces more accurate object geometry, better spatial layouts, and higher-fidelity textures compared to all baselines.

\FloatBarrier
\subsection{Additional Quantitative Results}

\methodname naturally extends to the single-image (image-to-3D) setting by treating one frame as a one-frame video; OVOW-3D-Scene-Bench in the main paper (Tab.~\ref{tab:quant_static}) reports exactly this comparison against all six baselines.
There, \methodname obtains the best Scene-IoU-OBB and Object-IoU and the lowest photometric and semantic error among all baselines; only VIGA attains a higher (axis-aligned) Scene-IoU-AABB.
Its strong layout accuracy in the single-image setting stems from the physics-grounded assembly stage, which enforces ground contact and inter-object support regardless of whether temporal information is available, while amodal inpainting keeps occluded-object geometry faithful.
The generative baselines (e.g., SceneGen and TabletopGen) place objects plausibly from their scene-generation priors but recover less accurate per-object geometry, whereas the segmentation-based SAM3D lacks explicit spatial-layout reasoning.

\FloatBarrier
\section{Examples of Simulation and Editing}
\setcounter{table}{0}
\setcounter{figure}{0}
\label{sec:supp_sim_edit}

A distinctive advantage of \methodname is that the reconstructed 4D scenes are \emph{directly usable} in downstream applications without any post-processing. We demonstrate two representative use cases: physics simulation and scene editing.

\begin{figure}[htbp]
    \centering
    \includegraphics[width=1\linewidth]{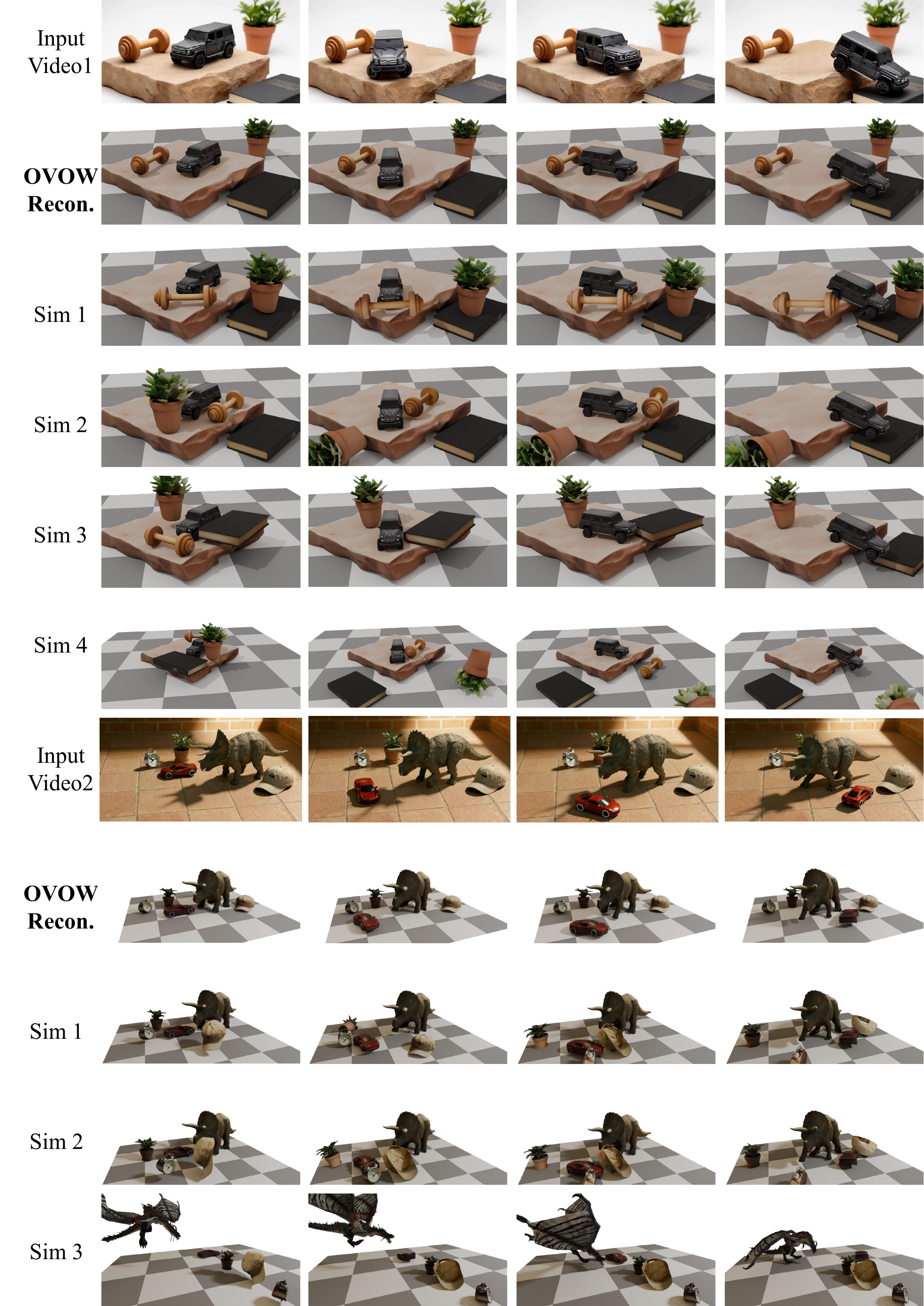}
    \caption{\textbf{Examples of downstream simulation and editing.} \emph{Top}: physics simulation in Blender (collision detection, gravity, and rigid- and soft-body dynamics) on the reconstructed scenes. \emph{Bottom}: scene editing operations in Blender, including object rearrangement, deletion, duplication, and re-texturing, enabled by instance-level mesh decomposition.}
    \label{fig:sim_edit}
\end{figure}

\noindent\textbf{Physics Simulation.}
Because the reconstructed assets are watertight, instance-separated meshes, they are directly usable for physics simulation, which we demonstrate in Blender's physics engine. Fig.~\ref{fig:sim_edit} (top) shows rigid-body objects undergoing realistic collision and stacking dynamics, while deformable objects exhibit plausible soft-body deformations under gravity and contact forces. The same geometry can be exported (e.g., in URDF format) for mainstream simulators such as MuJoCo~\cite{mujoco}, Isaac Gym~\cite{isaacgym}, and PyBullet~\cite{pybullet}. The downstream physics simulation discussed in the main paper confirms that the reconstructed assets are physically coherent and remain stable under gravity.

\noindent\textbf{Scene Editing.}
The instance-level decomposition enables intuitive scene editing in standard 3D tools such as Blender. Users can: (1) freely rearrange individual object instances by modifying their 6-DoF poses; (2) delete or duplicate specific objects; (3) replace object meshes while preserving the original motion trajectories; (4) retexture objects independently; and (5) composite reconstructed objects into new scenes. Fig.~\ref{fig:sim_edit} (bottom) illustrates several editing operations applied to reconstructed scenes, demonstrating the practical value of instance-level 4D mesh representations.

\FloatBarrier
\section{Failure Cases}
\setcounter{table}{0}
\setcounter{figure}{0}
\label{sec:supp_failures}

We identify two primary failure modes of \methodname:

\noindent\textbf{Failure Mode 1: Scenes with Many Objects.}
When the scene contains a large number of objects (typically $>$10), the VLM-guided scene decomposition may fail to correctly identify and segment all instances, particularly when objects are small or heavily overlapping. This leads to missing instances or incorrect motion category assignments. Furthermore, with many objects, the scene-level point cloud from VGGT~\cite{vggt} becomes less accurate for individual objects due to the complexity of the scene geometry, making it difficult to recover correct metric scales and to place instances at their correct positions during physics-grounded assembly.

\noindent\textbf{Failure Mode 2: Extreme Deformations.}
Not all object motions can be faithfully captured by our deformable mesh representation. Specifically, when an object undergoes \emph{topological changes} during the video, such as a person pulling an object out of a bag, an object breaking apart, or a fluid being poured, the assumption of consistent mesh topology across frames is violated. In these cases, the video-to-4D model (Motion324~\cite{motion324}) produces corrupted mesh sequences with severe geometric artifacts. Similarly, very large deformations that significantly alter the object's overall shape (e.g., a cloth being unfolded from a compact bundle) can exceed the capacity of the per-vertex displacement representation.

\FloatBarrier
\section{Limitations and Future Work}
\setcounter{table}{0}
\setcounter{figure}{0}
\label{sec:supp_limitations}

Beyond the failure cases discussed above, we acknowledge several broader limitations and outline directions for future work:

\begin{enumerate}[leftmargin=1.5em,itemsep=3pt]
    \item \textbf{Foundation Model Dependencies.}
    As a training-free pipeline, \methodname inherits the failure modes of all underlying foundation models. VLM misclassification of motion categories (e.g., labeling a slowly moving object as ``static'') propagates to downstream stages. Feed-forward 3D generators (Hi3DGen~\cite{ye2025hi3dgen}) may produce low-quality meshes for highly occluded ($>$80\%) or rare object categories. Depth estimation errors from VGGT~\cite{vggt} can lead to incorrect metric scale recovery. Future work could incorporate confidence-aware fusion or self-consistency checks to mitigate these cascading errors.

    \item \textbf{Limited Physical Interaction Modeling.}
    The current physics-grounded assembly handles gravity-aligned contact and simple stacking configurations but does not model complex physical interactions such as articulated joints (e.g., a door hinge), deformable-deformable contact (e.g., cloth draping over another cloth), or friction-dependent arrangements. Incorporating learned physical priors or differentiable physics simulation could enable richer contact modeling.

    \item \textbf{Challenging Visual Conditions.}
    Thin or highly reflective objects (e.g., glass, mirrors), transparent objects, and objects with minimal texture challenge both the segmentation and depth estimation stages. Motion blur and severe lighting changes can also degrade pose tracking accuracy.

    \item \textbf{Static Camera Assumption.}
    While our pipeline handles moderate camera motion through VGGT-based camera estimation, very large camera motions (e.g., 360$^{\circ}$ walkaround) or rapid camera shaking can degrade the quality of the scene point cloud and depth maps, leading to inaccurate metric scale recovery.

    \item \textbf{Background Reconstruction.}
    \methodname focuses on foreground object instances and does not reconstruct background elements (walls, floors, outdoor terrain). Future extensions could integrate background reconstruction methods to produce complete scene representations.

\end{enumerate}

\noindent We believe that addressing these limitations will further expand the applicability of structured Video-to-4D reconstruction for embodied AI, robotics, and interactive content creation.

\end{document}